\documentclass[10pt,journal,compsoc]{IEEEtran}
\usepackage{graphicx}
\usepackage{amsmath,amssymb}
\usepackage{color, colortbl}
\usepackage{mathrsfs}
\usepackage{algorithm}
\usepackage{algpseudocode}
\usepackage{multirow}
\usepackage{caption}
\usepackage{subcaption}
\usepackage{makecell}

\newlength\savedwidth
\newcommand\Thline{\noalign{\global\savedwidth\arrayrulewidth\global\arrayrulewidth 0.8pt}\hline\noalign{\global\arrayrulewidth\savedwidth}}

\usepackage{booktabs}
\usepackage{diagbox}

\usepackage{xcolor,colortbl}
\definecolor{Gray}{gray}{0.85}
\definecolor{LightCyan}{rgb}{0.88,1,1}
\newcolumntype{a}{>{\columncolor{Gray}}c}

\ifCLASSOPTIONcompsoc
  \usepackage[nocompress]{cite}
\else
  \usepackage{cite}
\fi
\ifCLASSINFOpdf
\else
\fi
\usepackage{url}

\begin{document}
%
\title{TCFormer: Visual Recognition via Token Clustering Transformer}
%
%
%
%

\author{Wang Zeng, Sheng Jin, Lumin Xu, Wentao Liu, Chen Qian, Wanli Ouyang,~\IEEEmembership{Senior Member,~IEEE,} \\ Ping Luo,~\IEEEmembership{Member,~IEEE,} Xiaogang Wang,~\IEEEmembership{Senior Member,~IEEE}
\IEEEcompsocitemizethanks{\IEEEcompsocthanksitem Wang Zeng and Lumin Xu  are with The Chinese University of Hong Kong, Hong Kong SAR, China.\protect\\
E-mail: \{zengwang, luminxu\}@link.cuhk.edu.hk

\IEEEcompsocthanksitem Sheng Jin and Ping Luo are with The University of Hong Kong, Hong Kong SAR, China.\protect\\
E-mail: \{js20@connect, pluo@cs\}.hku.hk

\IEEEcompsocthanksitem Wentao Liu and Chen Qian are with SenseTime Research and Tetras.AI.\protect\\
E-mail: \{liuwentao, qianchen\}@tetras.ai

\IEEEcompsocthanksitem Wanli Ouyang is with The University of Sydney and Shanghai AI Laboratory.\protect\\
E-mail: wanli.ouyang@sydney.edu.au

\IEEEcompsocthanksitem Xiaogang Wang are with with The Chinese University of Hong Kong, Hong Kong SAR, China and Centre for Perceptual and Interactive Intelligence Limited.\protect\\
E-mail: xgwang@ee.cuhk.edu.hk

} 
\thanks{Manuscript received XX XX, XXXX; revised XX XX, XXXX.}
}

%
%

\markboth{IEEE TRANSACTIONS ON PATTERN ANALYSIS AND MACHINE INTELLIGENCE}%
{Shell \MakeLowercase{\textit{et al.}}: Bare Demo of IEEEtran.cls for Computer Society Journals}
%



\IEEEtitleabstractindextext{%
\begin{abstract}

Transformers are widely used in computer vision areas and have achieved remarkable success. Most state-of-the-art approaches split images into regular grids and represent each grid region with a vision token. However, fixed token distribution disregards the semantic meaning of different image regions, resulting in sub-optimal performance. To address this issue, we propose the Token Clustering Transformer (TCFormer), which generates dynamic vision tokens based on semantic meaning. Our dynamic tokens possess two crucial characteristics: (1) Representing image regions with similar semantic meanings using the same vision token, even if those regions are not adjacent, and (2) concentrating on regions with valuable details and represent them using fine tokens. Through extensive experimentation across various applications, including image classification, human pose estimation, semantic segmentation, and object detection, we demonstrate the effectiveness of our TCFormer. The code and models for this work are available at \url{https://github.com/zengwang430521/TCFormer}.

\end{abstract}

}

\maketitle

\IEEEdisplaynontitleabstractindextext

%
\IEEEpeerreviewmaketitle

\IEEEraisesectionheading{\section{Introduction}\label{sec:introduction}}

%
%
%
%
\begin{figure}[tb]
	\centering
	\includegraphics[width=0.48\textwidth]{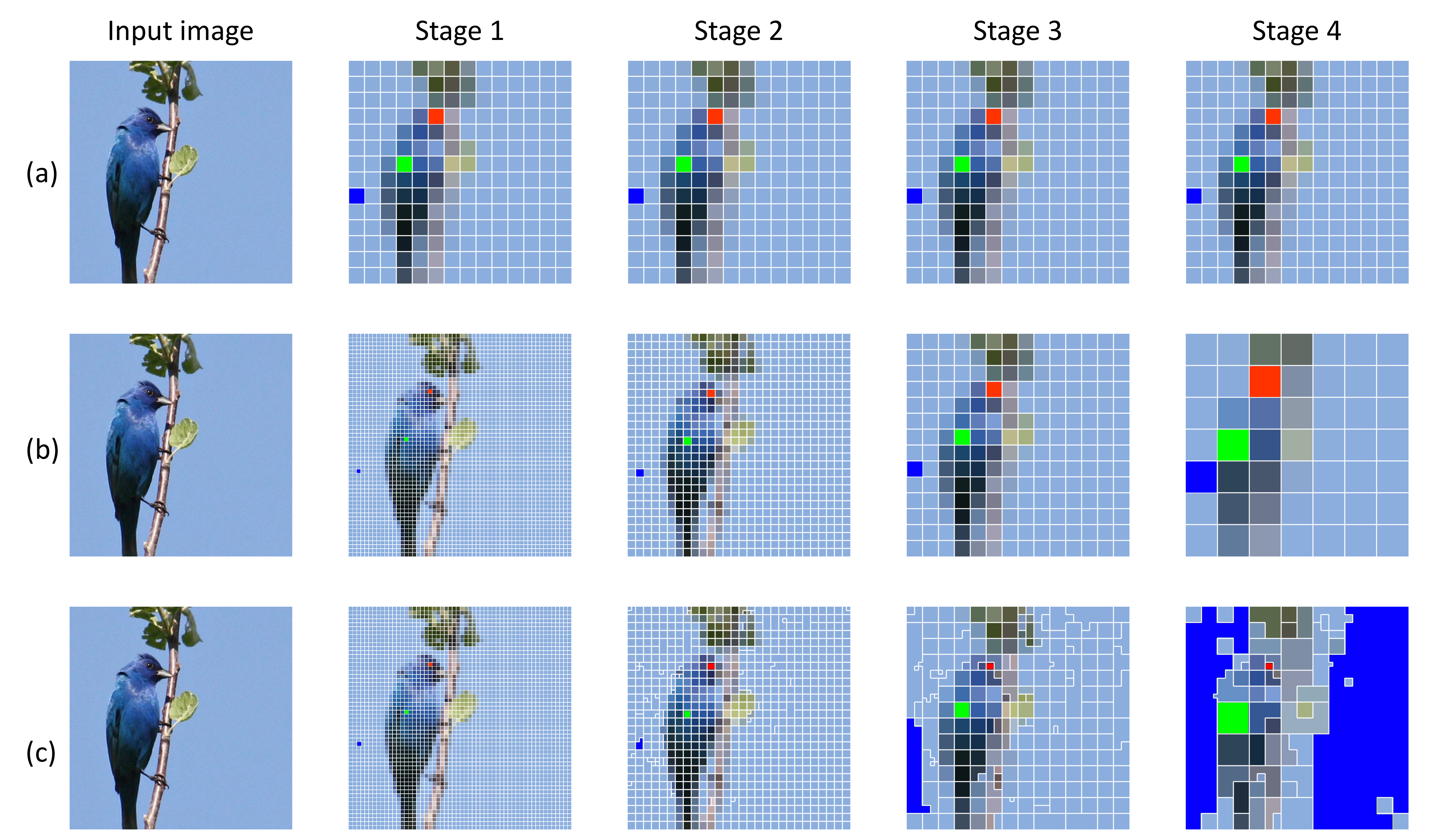}
	\caption{
        Comparisons of different vision token distribution. Image regions with the same color are represented by the same vision token. 
        Both prior isotropic(a) and pyramid(b) vision transformers treat all regions equally and disregard the differences in semantic meaning.
        In contrast, our TCFormer(c) generates dynamic vision tokens with flexible shapes and sizes based on the semantic meaning. 
        For the background regions, a single token (in blue) represents a large region, while for informative regions, more tokens (in green and red) are assigned. 
        For image details, tokens with fine spatial sizes are employed (in red).
        }
	\label{fig:introduction}
\end{figure}

\IEEEPARstart{V}{ision}	transformers have garnered state-of-the-art performance across a broad spectrum of tasks, including image classification~\cite{dosovitskiy2020image,touvron2021training,touvron2021going,wu2021cvt}, object detection~\cite{carion2020end,zhu2020deformable,zhang2022dino,liu2021swin}, semantic segmentation~\cite{xie2021segformer,cheng2022masked,strudel2021segmenter,chen2022vision}, pose estimation~\cite{yuan2021hrformer,xu2022vitpose,zeng2022not} and so on. 
The capacity for long-range attention between image patches affords vision transformers the ability to more effectively model relationships between image regions, thereby allowing for the acquisition of more robust representations in comparison to conventional convolutional neural networks.

Vision transformers approach images as sequences of feature vectors, known as vision tokens, each representing a specific image region. These tokens are then processed through transformer blocks.
%
While the architecture~\cite{dosovitskiy2020image,wang2022pvt, liu2021swin}, block structure~\cite{wu2021cvt}, and attention mechanism~\cite{wang2022pvt, liu2021swin, meng2022adavit} of vision transformers have undergone extensive investigation, the generation of vision tokens remains an under-explored area. 
To date, the majority of previous studies have relied on grid-based vision tokens.
%
As depicted in Fig.~\ref{fig:introduction}, isotropic vision transformers [1, 2] partition images using a fixed grid and consider each grid patch as a vision token. This design is straightforward and efficient, but lacks the ability to account for image features at varying scales. To address this issue, vision transformers with pyramid architectures [4, 8, 16, 18] employ different grid resolutions across different stages. 
%
While grid-based token generation has demonstrated impressive results across a broad range of tasks, it disregards the semantic meaning of images and treats all regions as equivalent, thus yielding sub-optimal results.

To tackle this challenge, we introduce a novel vision transformer, the Token Clustering Transformer (TCFormer), that dynamically generates vision tokens based on the semantic meaning of images. 
TCFormer incorporates a widely-used pyramid architecture. As illustrated in Fig.~\ref{fig:introduction}(c), we initiate from a high-resolution feature map and consider each pixel in the feature map as a vision token. Subsequently, we progressively merge these tokens through token feature clustering to generate dynamic tokens for the subsequent stage. To avoid too large complexity, we perform local clustering in early stages and execute global clustering in the final stage.
Unlike previous hierarchical vision transformers, TCFormer integrates tokens based on the similarity of their features, rather than their spatial positions, during the down-sampling process.

Our dynamic tokens have three key advantages. Firstly, they are better aligned with the objects in images, which enhances the learning of object relationships. Secondly, by allotting a larger number of tokens to valuable image regions, our dynamic tokens can concentrate on important regions and learn a more comprehensive representation of the image. Finally, the dynamic tokens can more effectively capture detailed information by representing image details using fine tokens.

Previous studies~\cite{sun2019deep,lin2017feature,zhu2020deformable} have demonstrated the advantages of multi-scale feature aggregation in a variety of tasks. For conventional grid-based vision tokens, prior works~\cite{wang2021pyramid,liu2021swin} first transform them into feature maps and perform feature aggregation using convolutional neural networks. However, converting our dynamic tokens into feature maps is not a straightforward task. The detailed information present in our dynamic tokens would be lost if transformed into low-resolution feature maps, while transforming them into high-resolution feature maps would result in an excessive computational burden.

%
To address this issue, we propose a Multi-stage Token Aggregation (MTA) module based on transformers. The MTA module regards the tokens from the last stage as initial tokens and gradually aggregates features from previous stages. At each step, the MTA module up-samples the vision tokens and merges them with token features from the preceding stage. Finally, the vision tokens at each step are combined to create a feature pyramid for subsequent processing.
To fully leverage the benefits of our dynamic token, we propose a new transformer block, that guides the attention process with clustering results. We incorporate the new transformer block in the MTA module to form the novel Clustering Reduction MTA (CR-MTA) module. 
The CR-MTA module achieves token feature aggregation in the form of vision tokens, preserving detailed information while maintaining an acceptable level of complexity.

The main contributions of this work can be summarized as follows:
 \begin{enumerate}
    \item[$\bullet$]
    We introduce a novel vision transformer, named TCFormer, which employs token feature clustering to generate dynamic vision tokens.

    \item[$\bullet$] 
    We propose a local CTM module that alleviates the excessive burden of token clustering by clustering tokens locally in early stages.
    
    \item[$\bullet$] 
    We propose a transformer based multi-scale feature aggregation module, termed MTA, that effectively and efficiently fuses multi-scale features in the form of vision tokens. 

    \item[$\bullet$] 
    We further improve the MTA module to CR-MTA module by incorporating a new transformer block guiding the attention process with clustering results.

    \item[$\bullet$]
    Extensive experiments on various computer vision tasks, including image classification, human pose estimation, semantic segmentation, and object detection, demonstrate the superiority of our proposed TCFormer over its counterparts.
\end{enumerate}

This work builds upon a preliminary version~\cite{zeng2022not} by incorporating the following enhancements:
1) We propose a local token clustering method that significantly reduces computation costs without sacrificing performance. The local clustering method enables our TCFormer to process high-resolution images more efficiently.
2) We propose CR-MTA to improve the original MTA module by guiding the attention process with clustering results. This enhancement fully utilizes the advantages of our dynamic vision tokens and improves the learning of object relationships.
3) We extend TCFormer to more tasks, such as object detection and semantic segmentation, to substantiate the versatility of our TCFormer.

\begin{figure*}[tb]
	\centering
	\includegraphics[width=0.96\textwidth]{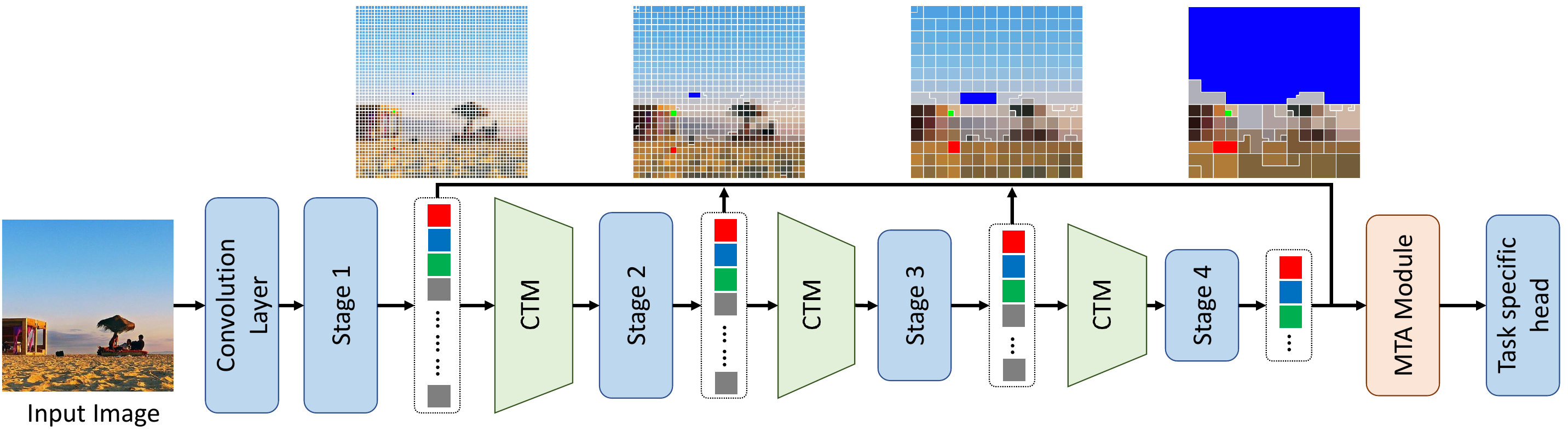}
	\caption{
        Architecture of our Token Clustering Transformer (TCFormer). TCFormer adopts a widely utilized pyramid structure and consists of four stages. The vision tokens in the initial stage are generated from the pixels in a high-resolution feature map. Between consecutive stages, the Clustering-based Token Merge (CTM) module merges vision tokens to create dynamic tokens for the subsequent stage. The Multi-stage Token Aggregation (MTA) module integrates multi-scale token features in token format and outputs a token pyramid for further processing.
        }
	\label{fig:arch}
\end{figure*}

\section{Related Work} \label{sec:related_work}

\subsection{Vision Transformer}
Inspired by the success of transformers in natural language processing [22-24], transformers have been extended to the field of computer vision and have produced state-of-the-art results across a wide range of tasks~\cite{dosovitskiy2020image,zhang2022dino,cheng2022masked,xu2022vitpose,liu2021swin}. The majority of state-of-the-art vision transformers divide images into grid-based patches and represent each patch as a vision token. The vision token sequence is then processed through multiple stacked transformer blocks. Based on the scale of the vision tokens, vision transformers can be divided into two categories.

%
The first type employs an isotropic architecture and uses vision tokens with uniform scales throughout the whole process. ViT~\cite{dosovitskiy2020image} applies isotropic vision transformers to image classification, predicting the classification score by processing the vision tokens along with a classification token. DeiT~\cite{touvron2021training} further introduces a distillation approach based on the classification token. MAE~\cite{he2022masked} expands ViT to a self-supervised learner by reconstructing randomly masked image patches. ViTPose~\cite{xu2022vitpose} leverages the ViT structure for pose estimation and achieves state-of-the-art performance.

The second type adopts pyramid structure and generates vision tokens with multiple scales.
%
The second type adopts a pyramid structure and generates vision tokens with multiple scales. CvT~\cite{wu2021cvt} extends ViT by incorporating convolutional layers into transformer blocks to construct multi-scale vision tokens. PVT~\cite{wang2021pyramid} introduces a vision transformer with a pyramid structure for dense prediction, using high-resolution grid-splitting in early stages and reducing the grid resolution after each stage. Swin~\cite{liu2021swin} has a similar structure as PVT, but employs shift window-based attention in its transformer blocks. Both PVT and Swin have achieved outstanding results in various tasks, such as object detection [7, 16] and semantic segmentation [10, 26].

%
Our TCFormer belongs to the second type and generates vision tokens with multiple scales. Unlike conventional models, the vision tokens in TCFormer are not restricted by a fixed grid structure. Instead, TCFormer generates dynamic tokens with adaptable shapes and sizes through token feature clustering, reflecting the semantic meaning of images and focusing on significant regions. TCFormer also captures image details with fine tokens. The dynamic vision tokens are beneficial for learning object relationships and capturing detailed information.

\subsection{Dynamic Vision Token}
%
Fixed grid-based vision tokens disregard the semantic meaning of images and treat all regions equally, which is sub-optimal. To address this issue, increasing attention is being paid to dynamic vision tokens. The key to generating dynamic vision tokens is to differentiate informative and uninformative image regions and focus on the informative regions.

%
Token pruning is a commonly used technique for generating dynamic tokens, which eliminates uninformative tokens to decrease computational complexity. DynamicViT~\cite{rao2021dynamicvit} and AdaViT~\cite{meng2022adavit} predict scores for each vision token and retain only the informative tokens with high scores. PnP-DETR~\cite{wang2021pnp} also selects informative tokens based on predicted scores. However, tokens with low scores are represented by coarse feature vectors. Evo-ViT~\cite{xu2022evo} distinguishes informative tokens based on the attention weights of the classification token and represents uninformative tokens with a single representative token.

Token pruning methods aim to reduce the computational cost of background image regions, while other methods aim to enhance the learning of image features. DVT~\cite{wang2021not} determines the token resolution based on the classification difficulty of the input image, enabling finer tokens to represent complex images. PS-ViT~\cite{yue2021vision} gradually adjusts the center of image patches to concentrate the vision tokens on informative regions and improve image features.

Compared to prior methods, the dynamic token generation in our TCFormer is more flexible. Previous methods are still confined to grid-based tokens, with token pruning adjusting the number of grid patches, DVT adjusting the grid scale, and PS-ViT adjusting the grid patch centers. Conversely, our TCFormer is not restricted to grid-based image patches and is adaptable in both token shape and size.
First, TCFormer generates tokens with flexible shapes. Our dynamic vision tokens divide images based on semantic meaning and are not restricted by spatial relationships, allowing even non-adjacent areas to be represented by a single token. This flexibility in token shapes enhances alignment with objects in images, improving the learning of object relationships. Secondly, TCFormer dynamically adjusts token density, allocating more tokens to informative regions to learn more representative image features. Thirdly, TCFormer employs tokens of varying scales for different regions, enabling the capture of detailed information through the representation of image details with fine tokens.

Recently, there have also been clustering-based methods for dynamic token generation. \cite{bolya2022token} generates dynamic tokens by gradually merging the most similar token pairs, and \cite{liang2022expediting} proposes the use of iterative local clustering for token merging. These works share the same idea as our TCFormer, which suggests that image regions should be divided according to semantic meaning rather than spatial location. While both ~\cite{bolya2022token} and \cite{liang2022expediting} aim to reduce computational complexity by decreasing the number of tokens, in contrast, our goal is to enhance image features through dynamic tokens.


%




\section{Token Clustering Transformer} \label{sec:tcformer}
The overall architecture of Token Clustering Transformer (TCFormer) is shown in Fig.~\ref{fig:arch}. TCFormer adopts a popular pyramid structure, consisting of four stages. Each stage is composed by several stacked transformer blocks. A Clustering-based Token Merge (CTM) module is interspersed between adjacent stages to dynamically merge tokens and enable each stage to process tokens at varying scales. The vision tokens in the first stage are initialized from a high-resolution feature map, with each pixel in the feature map regarded as a token. After the final stage, a Multi-stage Token Aggregation (MTA) module integrates multi-scale features in the form of vision tokens and outputs a pyramid of image features for task-specific processing. 
We will introduce the preliminary version (TCFormerV1) in Section~\ref{sec:tcformer:v1} and the new version (TCFormerV2) in Section~\ref{sec:tcformer:v2}.

\subsection{TCFormerV1}
\label{sec:tcformer:v1}

\begin{figure}[tb]
	\centering
	\includegraphics[width=0.48\textwidth]{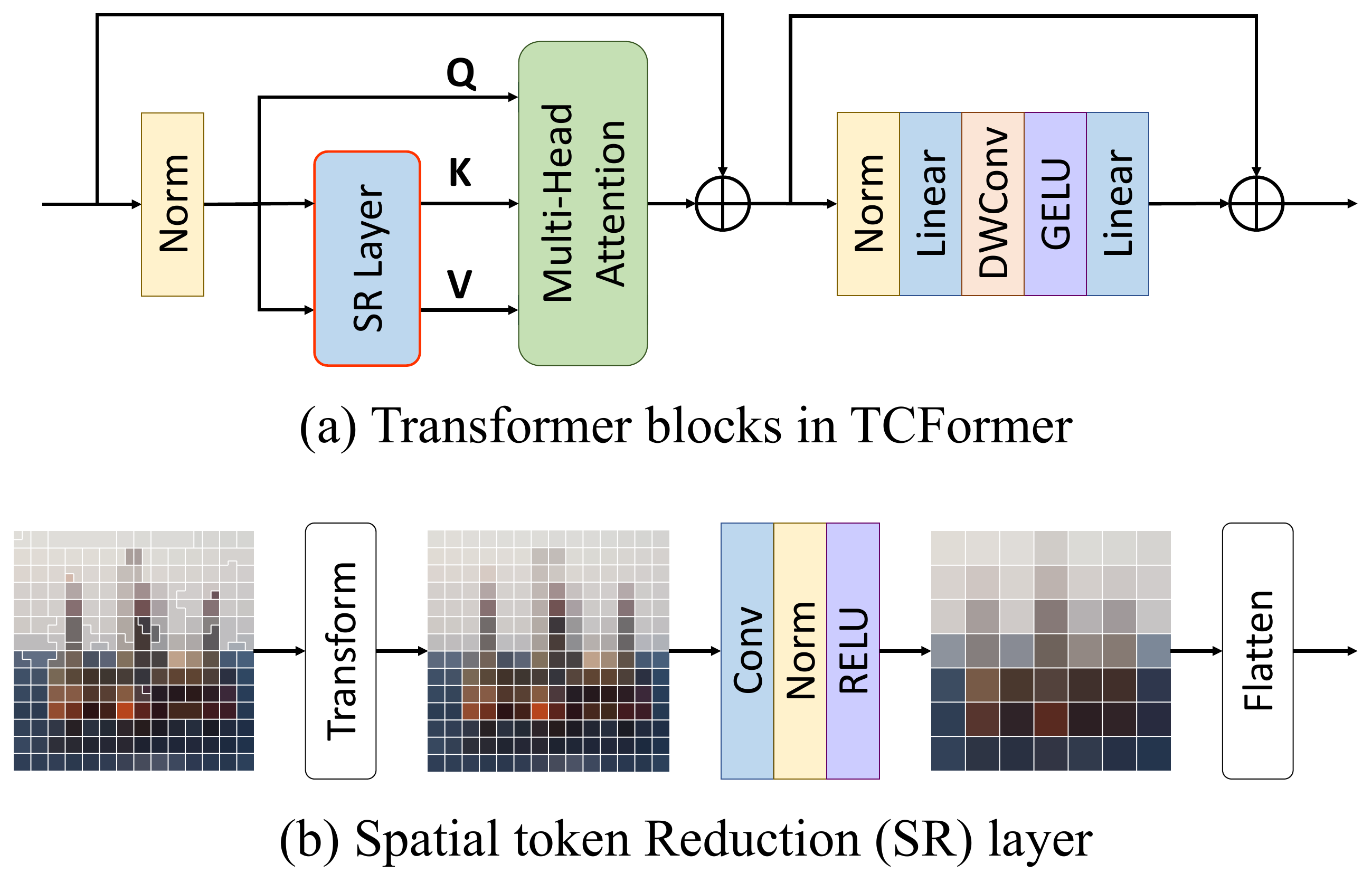}
        \caption{
        (a) Structure of the transformer blocks in TCFormerV1. Before the attention module, a token reduction layer is inserted to reduce the computation complexity. After the attention module, a depth-wise convolutional layer is included to extract local information.
        (b) The Spatial Token Reduction (SR) layer converts dynamic tokens into a feature map, which is subsequently compressed and flattened into key and value tokens.
        }
	\label{fig:SR_block}
\end{figure}

\subsubsection{Transformer Block}
\label{sec:tcformer:block}
Fig.~\ref{fig:SR_block} illustrates the structure of the transformer blocks in TCFormerV1. Due to the typically large number of tokens in vision transformers, the global attention processes in traditional transformer blocks result in an unacceptable computational cost. To address this issue, we introduce a token reduction layer before the attention process to reduce the number of tokens.
As depicted in Fig.~\ref{fig:SR_block}(b), the Spatial Token Reduction (SR) layer converts vision tokens into a feature map and then reduces the resolution of the feature map using a convolutional layer. The down-sampled feature map is subsequently flattened to form the key and value tokens, resulting in a reduced number of tokens.
Following the attention process, we incorporate a depth-wise convolutional layer to capture local information, which has been demonstrated to be beneficial in previous works such as \cite{wu2021cvt,li2021localvit,yuan2021hrformer}.

\subsubsection{Clustering-based Token Merge (CTM) Module}
\label{sec:tcformer:ctm}
\begin{figure}[tb]
	\centering
	\includegraphics[width=0.48\textwidth]{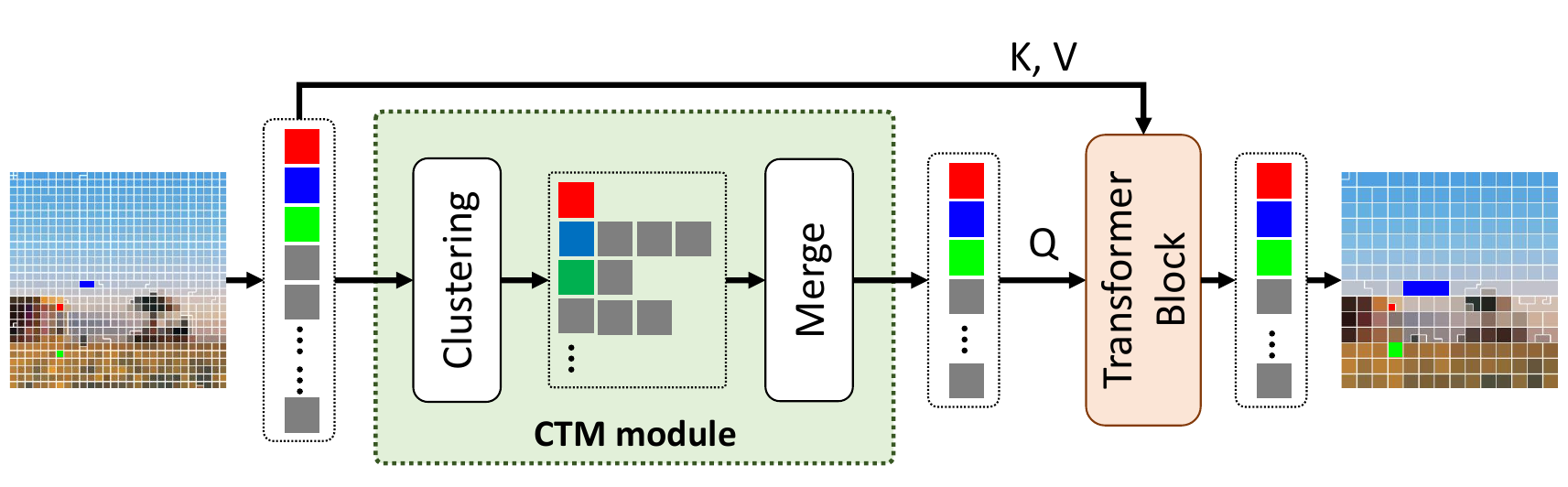}
        \caption{
        Illustration of the dynamic vision token generation process. The Clustering-based Token Merge (CTM) module first groups the input tokens into several clusters and then merges the tokens in the same cluster into a single token via weighted feature averaging. After the CTM module, the merged tokens and the original tokens are input into a transformer block for better feature aggregation.
        }
	\label{fig:ctm}
\end{figure}

As shown in Fig.~\ref{fig:ctm}, the Clustering-based Token Merge (CTM) module of our TCFormer is comprised of a token clustering and merging process. Given the vision tokens from the prior stage, the CTM module first groups the tokens into clusters through the application of a clustering algorithm to the token features, and then merges the tokens within the same cluster to generate new vision tokens for the subsequent stage.

\textbf{Token Clustering.}
For the token clustering process, we utilize a variation of the density peaks clustering algorithm based on k-nearest neighbors (DPC-kNN)~\cite{du2016study}, due to its simplicity and parallelization capabilities.
Given a set of vision tokens $X$, we first calculate the distance between each token and the other tokens. Then, we estimate the local density $\rho$ from the distances between the token and its k-nearest neighbors:
\begin{equation}
\rho_{i}=\exp \left(-\frac{1}{k} \sum_{x_{j} \in \mathrm{kNN}\left({x}_{i}\right)} 
\left\|x_i-x_j\right\|_{2}^{2},
\right),
\label{eq:rho}
\end{equation}
where $\mathrm{kNN}\left({x}_{i}\right)$ denotes the k-nearest neighbors of a token $i$. $x_i$ and $x_j$ are their corresponding token features.

Then for each token, we collect the distance between it and the tokens with higher local density, and use the minimal distance as the distance indicator. Tokens with large distance indicator tends to be local density peaks and are considered suitable candidates of clustering centers. For the token with highest local density, we set the maximum distance between it and other tokens as its distance indicator to make sure that it has the largest distance indicator.
\begin{equation}
\delta_{i}=\left\{\begin{array}{l}
\min _{j: \rho_{j}>\rho_{i}} \left\|x_i-x_j\right\|_{2}, \text { if } \exists j \text { s.t. } \rho_{j}>\rho_{i} \\
\max _{j} \left\|x_i-x_j\right\|_{2}, \text { otherwise }
\end{array}\right.
\label{eq:delta}
\end{equation}
where $\delta_{i}$ denotes the distance indicator and $\rho_{i}$ denotes the local density.

Finally, we determine the score of each token by multiplying its local density by its distance indicator, resulting in $\rho_i \times \delta_i$. Tokens with higher scores possess a greater likelihood of being cluster centers. The tokens with the highest scores are then selected as clustering centers, and the remaining tokens are allocated to the nearest center.

\textbf{Token Merging.}
The token merging process endeavors to produce a single representative token for each cluster. A straightforward approach would be to simply average the token features within the cluster. However, this ignores the varying significance of each token. Drawing inspiration from prior works~\cite{rao2021dynamicvit,meng2022adavit,wang2021pnp}, we predict the importance score $P$ of each token based on its features to reflect its significance and guide the averaging of token features using the predicted importance scores:
\begin{equation}
y_{i}=\frac {\sum_{j \in C_i} e^{p_j} x_j}
            {\sum_{j \in C_i} e^{p_j}} ,
\label{eq:average}
\end{equation}
where $C_i$ means the set of the $i$-th cluster, $x_j$ and $p_j$ are the original token feature and the corresponding importance score respectively. $y_i$ is the features of the merged token.

As depicted in Fig.~\ref{fig:ctm}, the original tokens and the merged tokens are input into the subsequent transformer block, with the merged tokens functioning as queries and the original tokens serving as keys and values. The transformer block aims to transfer features from the original tokens to the merged tokens, with the importance score $P$ incorporated into the attention mechanism to steer the feature transfer process.
\begin{equation}
\operatorname{Attention}(Q, K, V)=\operatorname{softmax}\left(\frac{Q K^{T}}{\sqrt{d_{k}}}+P\right) V,
\label{eq:attn}
\end{equation}
where $d_k$ is the channel dimension of the queries. 
For clarity, we omit the multi-head setting and the spatial reduction layer. By incorporating the token importance score into both the feature averaging and attention processes, we ensure that the crucial visual tokens have a greater impact on the output dynamic tokens.

\begin{figure}[tb]
	\centering
	\includegraphics[width=0.48\textwidth]{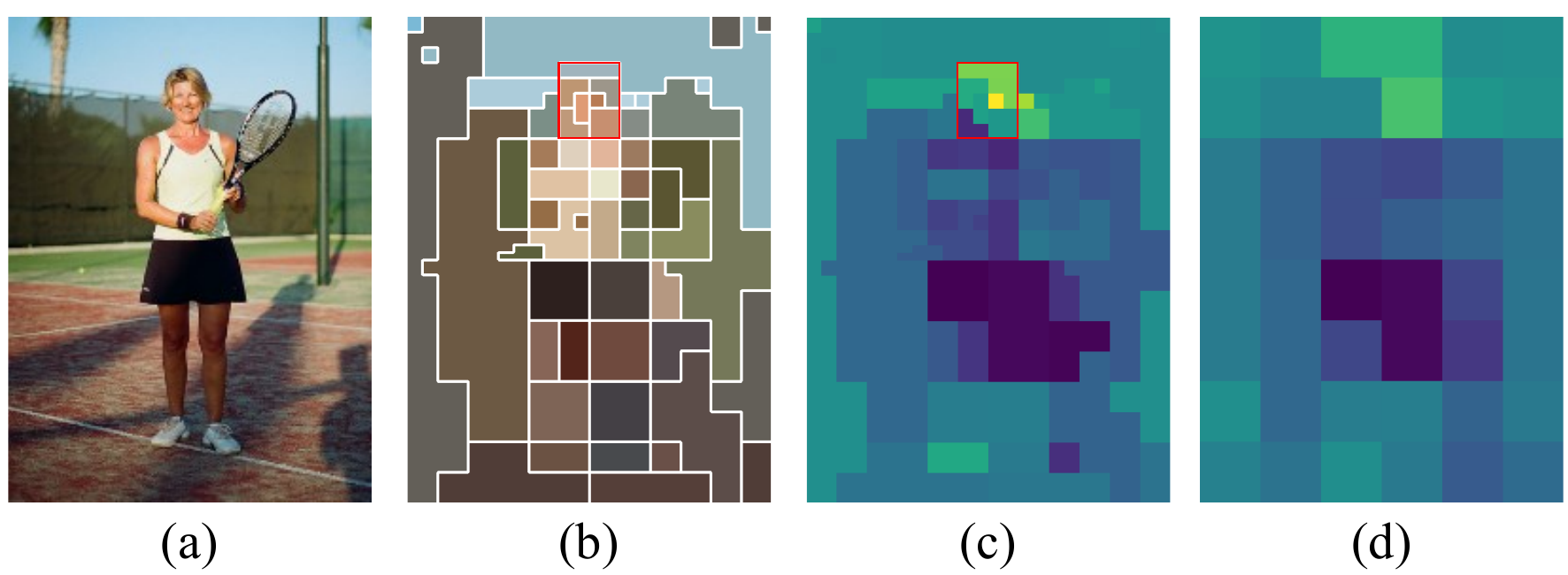}
        \caption{
        A typical example of the dynamic tokens produced by TCFormer. The input image is depicted in (a), and the dynamic tokens are presented in (b). The dynamic tokens can be converted into a high-resolution feature map (c), which retains the details but leads to large computational complexity, or a low-resolution feature map (d), which sacrifices the detailed information in the dynamic tokens.
        }
	\label{fig:mta_token}
\end{figure}

\begin{figure}[tb]
	\centering
	\includegraphics[width=0.47\textwidth]{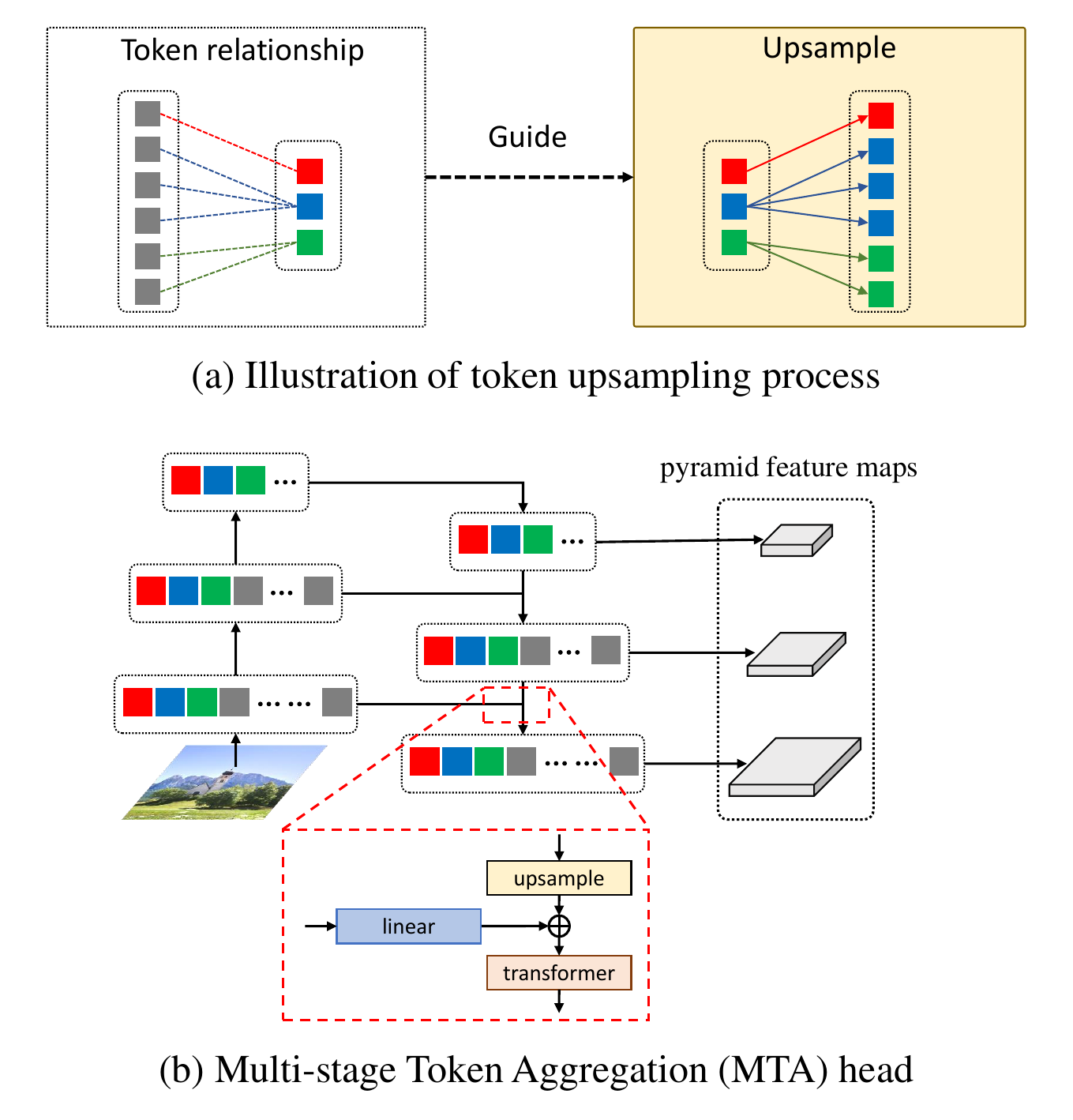}
        \caption{
        Illustration of the Multi-Stage Token Aggregation (MTA) Module. (a) During the token upsampling process, we use the recorded token relationship to copy the merged token features to the corresponding upsampled tokens. (b) The MTA module starts from the final stage and progressively accumulates features through sequential upsampling procedures and transformer blocks. The tokens at every stage are transformed to create a feature map pyramid for further processing.
        }
	\label{fig:mta}
        \vspace{-2mm}
\end{figure}

\subsubsection{Multi-stage Token Aggregation (MTA) Module}
\label{sec:tcformer:mta}
The integration of multi-scale features has been demonstrated to be beneficial for various tasks~\cite{sun2019deep,lin2017feature,zhu2020deformable}. Previous vision transformers~\cite{wang2021pyramid,liu2021swin,yuan2021hrformer} transform vision tokens into feature maps and aggregate multi-scale features using conventional convolutional neural networks.

However, CNN-based feature aggregation modules are not suitable for our dynamic vision tokens. As illustrated in Fig.~\ref{fig:mta_token}, TCFormer generates vision tokens without a grid structure and assigns fine tokens to image regions containing intricate details. Transforming our dynamic tokens into low-resolution feature maps, as done in previous works~\cite{sun2019deep,lin2017feature,zhu2020deformable}, results in a loss of detail. On the contrary, transforming the tokens into high-resolution feature maps retains the details but incurs unacceptable complexity and memory overhead. To alleviate this problem, we propose a new transformer-based Multi-stage Token Aggregation (MTA) module.

Following the popular FPN~\cite{lin2017feature}, our MTA module progressively aggregates features from deeper stages to shallower stages. 
We initially introduce an upsampling process for our dynamic vision token. 
As depicted in Fig.~\ref{fig:mta}(a), in the CTM modules, we group tokens into multiple clusters and consolidate the tokens within each cluster into a single token. The clustering results are recorded for the token upsampling process, wherein the features from the merged tokens are replicated to the corresponding upsampled tokens based on the recorded clustering results.

With the token upsampling process, we can systematically aggregate features on a stage-by-stage basis. Rather than converting vision tokens into feature maps, we introduce a transformer-based Multi-stage Token Aggregation (MTA) module designed to aggregate features in the token format. 
As illustrated in Fig.~\ref{fig:mta}(b), the visual tokens in the final stage serve as the initial tokens. At each step, the MTA module initially performs upsampling on the tokens, ensuring that the upsampled tokens share the same distribution as the tokens in the preceding stage. Subsequently, the MTA module integrates the token features from the preceding stage into the upsampled tokens, feeding the results into a transformer block. This iterative process continues until features from all stages have been effectively aggregated. Ultimately, the tokens at each step undergo transformation into pyramid feature maps for subsequent processing.

Diverging from FPN, which converts vision tokens into feature maps, our MTA module aggregates features in token format. This approach preserves details at every stage while avoiding the processing of high-resolution feature maps, achieving a harmonious balance between performance and efficiency.

\subsection{TCFormerV2}
\label{sec:tcformer:v2}
TCFormerV1 produces flexible dynamic tokens and achieves impressive results in human-centric tasks~\cite{zeng2022not}. However, the original CTM module is burdened with significant complexity for high-resolution input images.
Therefore, we further enhance TCFormerV1 by introducing a novel local CTM module (Section~\ref{Sec:LocalCTM}). 
To fully tap the potential of the dynamic tokens, we further propose a Clustering Reduction based Multi-stage Token Aggregation (CR-MTA) module (Section~\ref{sec:tcformer:cr_mta}).

\subsubsection{Local CTM}
\label{Sec:LocalCTM}
As outlined in Section~\ref{sec:tcformer:ctm}, the DPC-kNN algorithm~\cite{du2016study} employed in the original CTM modules involves computing distances between every pair of tokens.
 This process imposes a memory cost and computational complexity that are quadratic in relation to the number of tokens. Consequently, for high-resolution input images, the original CTM modules in early stages lead to an unacceptable level of complexity and memory usage.

\begin{figure}[tb]
	\centering
	\includegraphics[width=0.48\textwidth]{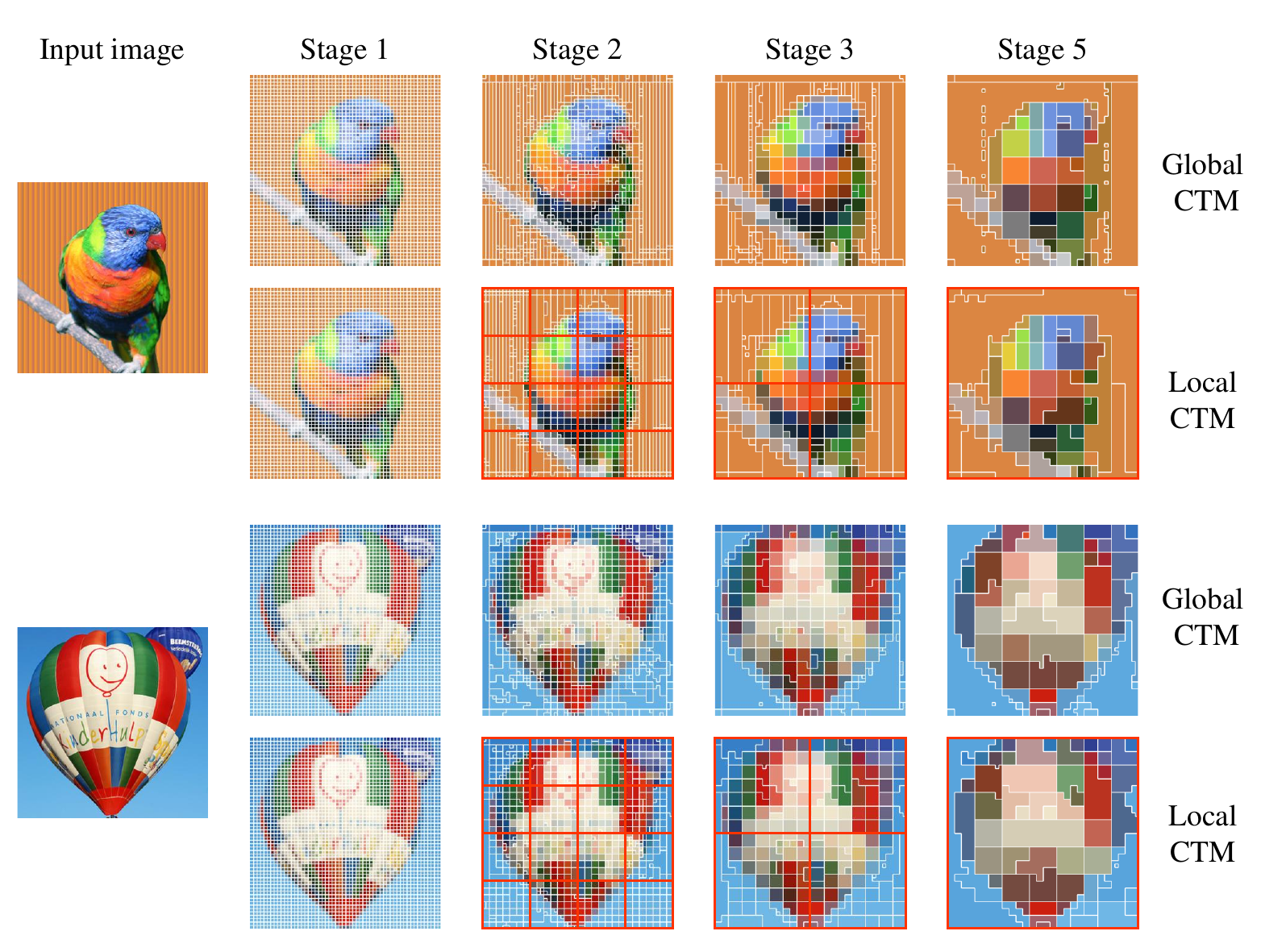}
        \caption{
        Examples of the dynamic tokens generated by global CTM modules and local CTM modules. Global and local CTM modules produce similar dynamic tokens.
        }
	\label{fig:local_ctm_token}
\end{figure}

As shown in Figure~\ref{fig:local_ctm_token}, the CTM module exhibits varying effects in different stages. In the early stages, it tends to merge vision tokens with their nearby tokens and aligns dynamic tokens with the edges of objects, for example the tree branches and the hot air balloon. 
In deep stages, the CTM module merges distant tokens based on high-level semantic meanings, exemplified by the wall and the sky region.
Leveraging this characteristic of the CTM module, we introduce a new module named the local CTM module,  which reduces complexity without compromising performance. For reference, the original CTM module is denoted as the global CTM module in subsequent sections.

\begin{figure}[tb]
	\centering
	\includegraphics[width=0.48\textwidth]{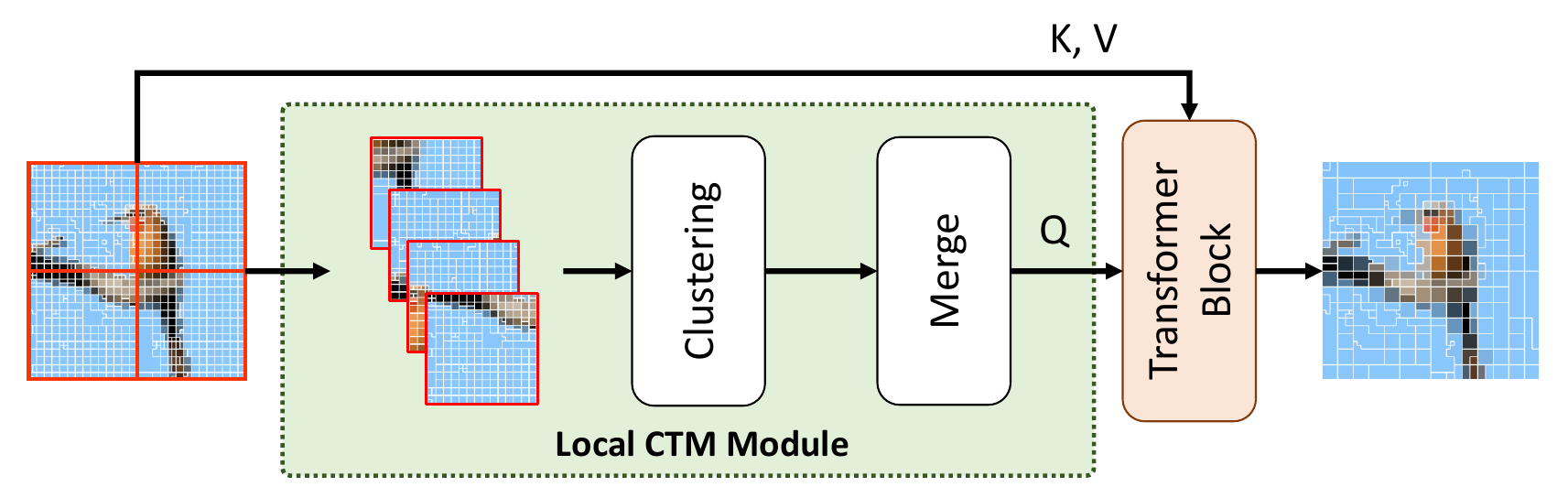}
        \caption{
        Illustration of the local CTM module. Local CTM module divides input tokens into several parts based on their spatial location and apply the clustering algorithm to each part individually.}
	\label{fig:local_ctm}
\end{figure}

As depicted in Figure~\ref{fig:local_ctm}, the local CTM module divides the dynamic tokens into multiple parts in early stages and performs a token clustering process for each part individually. This approach allows the output tokens to align with object edges while requiring computation of distances only between neighboring token pairs.
Assuming that we partition the input tokens into $P$ parts and the input tokens have a number of $N$ and a feature channel dimension of $C$, the computation complexity of the global clustering method is $N^{2}C$, while the complexity of the local clustering method is $N^{2}C/P$. The latter is reduced by a factor of $P$ compared to the former.

As we progress to deeper stages, we decrease the number of parts used in the local CTM module to facilitate the merging of more distant tokens. In the final stage, we reduce the part number to $1$, and the original global token clustering is performed in this module. The gradual reduction in part number allows image regions that are distant spatially but similar in semantic meanings to be represented by the same token.
We incorporate local CTM modules into our TCFormerV2. Specifically, we utilize a part number of $16$ and $4$ in the first two local CTM modules and employ a part number of $1$ in the last local CTM module.

In Figure~\ref{fig:local_ctm_token}, we contrast the dynamic tokens generated by global CTM modules and local CTM modules. 
In early stages, even with relatively large part numbers, the local CTM modules generate dynamic tokens that align well with object edges. 
In deep stages, the small part numbers in our local CTM module enable long-range token merging, generating dynamic tokens that align well with semantic meanings. 
In conclusion, our local CTM modules generate token distribution similar to the global CTM module but with significantly less complexity.

\subsubsection{Clustering Reduction MTA}
\label{sec:tcformer:cr_mta}
As depicted in Section~\ref{sec:tcformer:mta}, the MTA module employs transformer blocks for feature aggregation, ensuring the effective and efficient preservation of detailed information within our dynamic tokens.  
Nevertheless, the ordinary transformer blocks utilized in the original MTA module fail to fully leverage the advantages of our dynamic tokens and warrant further improvements.

\begin{figure}[tb]
	\centering
	\includegraphics[width=0.48\textwidth]{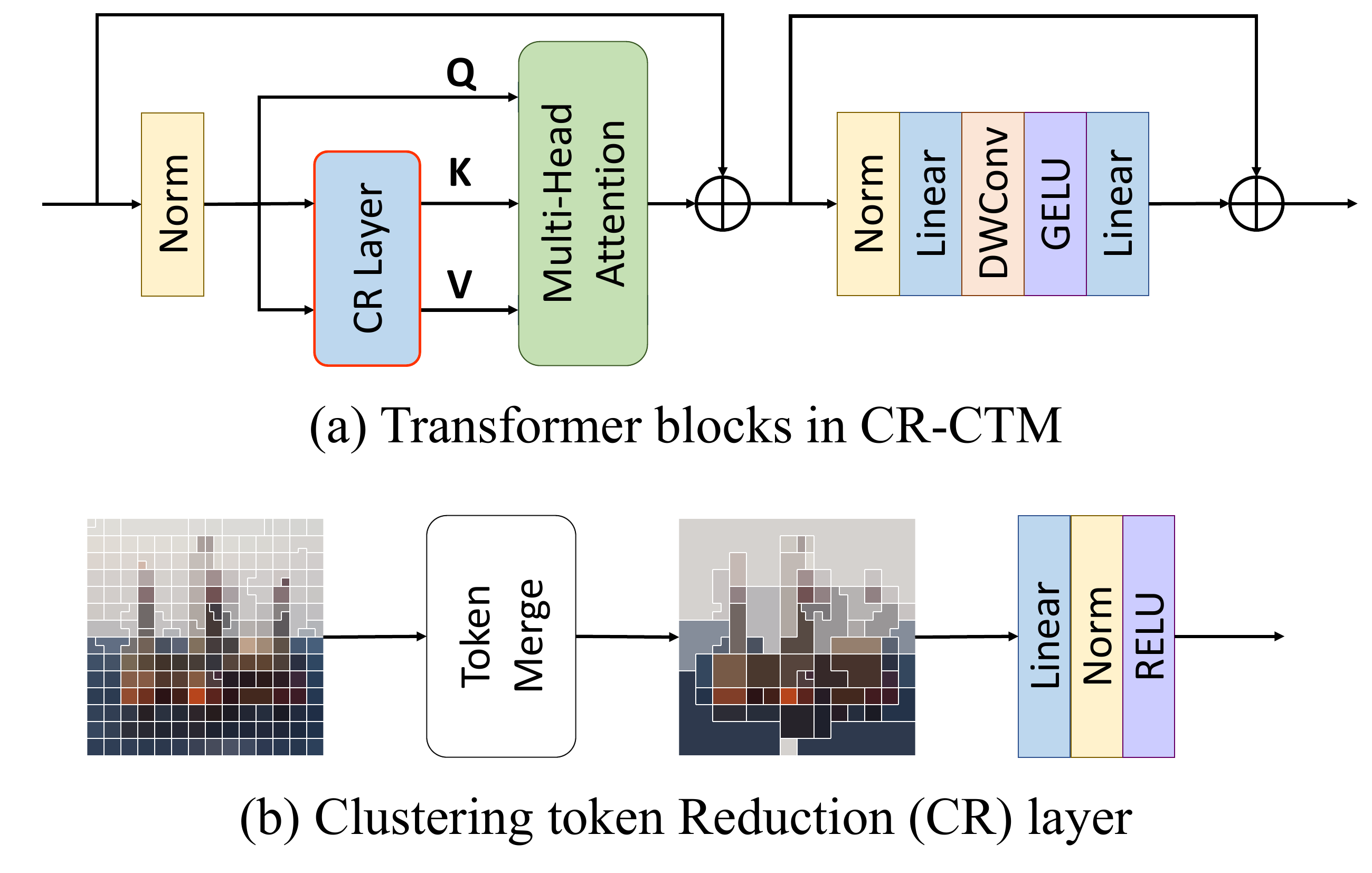}
        \caption{
        Illustration of the Spatial token Reduction (SR) layer (a) and the Clustering token Reduction (CR) layer (b).
        }
	\label{fig:CR_block}
 \end{figure}

\begin{figure}[tb]
	\centering
	\includegraphics[width=0.48\textwidth]{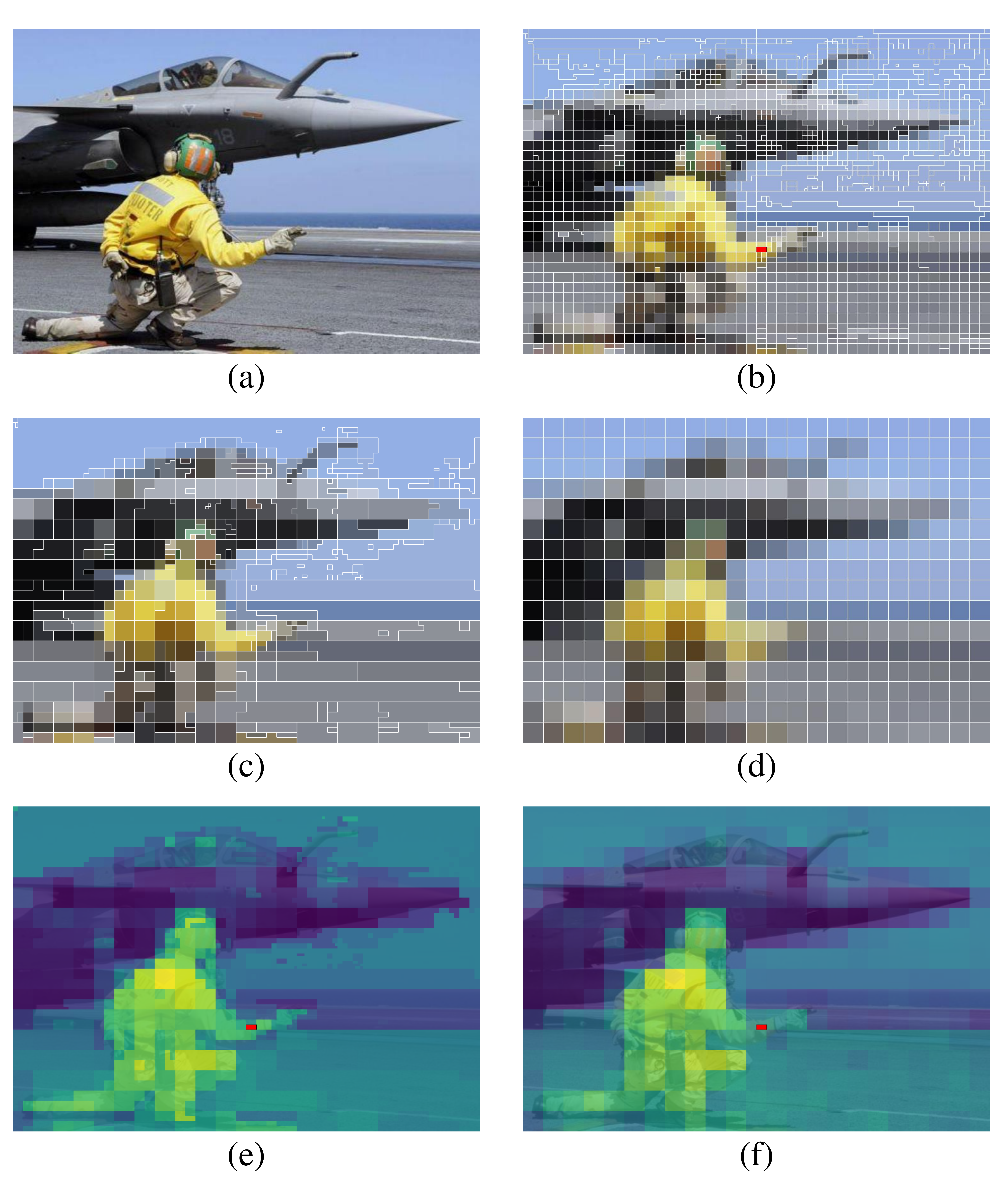}
        \caption{
        Illustration of the attention weight in the MTA modules for semantic segmentation task. A representative input image is displayed in (a), while the dynamic token generated by TCFormer is shown in (b). The output tokens of the CR module and SR module are visualized in (c) and (d) respectively.
        We choose a token on the human body (highlighted in red) and show its attention weight in the CR-MTA module (e) and in the SR-MTA module (f). The attention weight in the CR-MTA module aligns better with the objects in the image, which is beneficial to the learning of image features.
        }
	\label{fig:attn_vis}
\end{figure}

As described in Section~\ref{sec:tcformer:block}, our transformer blocks include a Spatial Token Reduction (SR) layer before the attention process to reduce the computational complexity. While the SR layer is simple and effective, it disrupts the alignment between vision tokens and objects in images. Specifically, as depicted in Figure~\ref{fig:attn_vis}(b), our dynamic tokens are well-aligned with the objects in the input image. This alignment promotes a clearer semantic understanding of vision tokens and simplifies the learning of object relationships. However, as shown in Figure~\ref{fig:CR_block}(d), the output tokens of the SR layers maintain a fixed grid-based distribution, resulting in the loss of alignment.

\textbf{CR Transformer Block.} 
To address this concern, we introduce a novel transformer block specifically tailored for our MTA module.
Our proposal begins with a Clustering Token Reduction (CR) layer.
As depicted in Fig.~\ref{fig:CR_block}(b), CR layers reduce the number of tokens by merging them based on the token clustering results generated by previous CTM modules.
The output of the CR layers shares the same distribution as the dynamic tokens in the final stage, thereby preserving the alignment between tokens and objects in images.
By reusing the clustering results from the CTM modules, CR layers do not impose additional computational burden and is more efficient than SR layers.
As shown in Fig.~\ref{fig:CR_block}(a), we build the new CR transformer block by replacing the SR layer in the ordinary transformer blocks with the new proposed CR layer.

\textbf{CR-MTA Module.}
We introduce a new CR-MTA module by replacing ordinary SR transformer blocks with CR transformer blocks and integrate the CR-MTA module into TCFormerV2. 
As the CR block relies on clustering results, we restrict its application to the MTA module and retain SR blocks in previous stages. 
Fig.~\ref{fig:attn_vis} presents a  visual comparison between the attention weights in the CR-MTA module and the original MTA module, denoted as SR-MTA module in subsequent sections. As shown in Fig.~\ref{fig:attn_vis}(c), the output tokens of the CR layer exhibit a commendable alignment between the person and the airplane in the input image. Therefore, as shown in Fig.~\ref{fig:attn_vis}(e), the attention weights in the CR-MTA module align with a sharp outline of the human body, including detailed parts such as fingers. In contrast, as shown in Figure~\ref{fig:CR_block}(f), the attention weights in the SR-MTA module only correspond to the rough human body outline.

\subsection{Comparison of TCFormer Variant}
The summary of the difference between TCFormerV1 and TCFormerV2 is presented below.

\noindent\textbf{TCFormerV1.} 
TCFormerV1 utilizes global CTM modules to generate dynamic tokens and a SR-MTA module to aggregate token features. This model is used in the preliminary paper~\cite{zeng2022not}.

\noindent\textbf{TCFormerV2.} 
TCFormerV2 is an improved version of TCFormerV1, accomplished by introducing local CTM modules and a CR-MTA module. Compared to TCFormerV1, TCFormerV2 is more efficient and can learn object relationships more effectively.

\section{Experiments}
In this section, we apply both TCFormerV1 and TCFormerV2 to  a variety of computer vision tasks, including image classification, human pose estimation, semantic segmentation, and object detection. 
Regarding image classification, both TCFormerV1 and TCFormerV2 outperforms their counterparts. However, TCFormerV2 exhibits lower complexity. 
In terms of human pose estimation, TCFormerV1 achieves impressive performance, and TCFormerV2 further improves the performance to a new state-of-the-art level. 
In semantic segmentation, TCFormerV1 surpasses grid-based vision transformers, but involves too much complexity. Conversely, TCFormerV2 achieves better performance with significantly less complexity. 
For object detection, we only apply TCFormerV2, as TCFormerV1 incurs unacceptable memory costs. TCFormerV2 demonstrates superiority over its counterparts, particularly for detecting small objects.
Detailed results are presented below.

\begin{table}[t]
    \centering
    \caption{Evaluation on ImageNet-1k \texttt{val}. 
    All results are obtained using the input size of $224\times 224$. 
    Throughput is measured using the GitHub repository of~\cite{rw2019timm} and a V100 GPU.}
    \small
    \label{tab:cls}
    \scalebox{0.82}{
     \begin{tabular}{l|c|c|c|c}
    \Thline
    \renewcommand{\arraystretch}{0.1}
	Method & \#Param & GFLOPs & \makecell[c]{Throughput \\ (image/
 s)} & Top-1 Acc  \\
	\hline
	ResNet18~\cite{he2016deep} & $11.7$ & $1.8$ & $3786.0$ & $69.8$  \\
	DeiT-Tiny/16~\cite{touvron2021training} & $5.7$ & $1.3$ & $2537.5$ & $72.2$ \\
	PVTv1-Tiny~\cite{wang2021pyramid} & $13.2$ & $1.9$ & $1410.1$ &$75.1$ \\
	PVTv2-B1~\cite{wang2022pvt}    & $13.1$ & $2.1$ & $1179.2$ & $78.7$ \\
        \rowcolor{Gray}
	TCFormerV1-Light & $14.2$ & $3.8$ & $185.9$ & $79.6$ \\
        \rowcolor{Gray}
        TCFormerV2-Tiny & $14.2$ & $2.5$ & $393.5$ & $79.5$ \\
	\hline
	ResNet50~\cite{he2016deep}   & $25.6$ & $4.1$ & $1163.4$ & $76.1$ \\
	ResNeXt50-32x4d~\cite{xie2017aggregated} & $25.0$ & $4.3$ & $862.9$ & $77.6$  \\
	RegNetY-4G~\cite{radosavovic2020designing} & $21.0$ & $4.0$ & $966.8$ & $80.0$ \\
	DeiT-Small/16~\cite{touvron2021training}  & $22.1$ & $4.6$ & $747.8$ & $79.9$ \\
	T2T-ViT$_t$-14~\cite{yuan2021tokens} & $22.0$ & $6.1$ & $-$ & $80.7$ \\
	PVTv1-Small~\cite{wang2021pyramid}  & $24.5$ & $3.8$ & $795.9$ & $79.8$ \\
	TNT-S~\cite{han2021transformer} & $23.8$ & $5.2$ & $410.0$ & $81.3$ \\
	Swin-T~\cite{liu2021swin} & $29.0$ & $4.5$ & $747.8$ & $81.3$ \\
	CvT-13~\cite{wu2021cvt} & $20.0$ & $4.5$ & $-$ & $81.6$ \\
	CoaT-Lite Small~\cite{xu2021co} & $20.0$ & $4.0$ & $616.5$ & $81.9$ \\
	Twins-SVT-S~\cite{chu2021twins} & $24.0$ & $2.8$ & $1006.21$ & $81.7$ \\
        iFormer-S~\cite{si2022inception} & $20.0$ & $4.8$ & $-$ & $83.4$ \\
	PVTv2-B2~\cite{wang2022pvt} & $25.4$ & $4.0$ & $674.1$ & $82.0$ \\
        \rowcolor{Gray}
        TCFormerV1 & $25.6$ & $5.9$ & $120.8$ & $82.4$ \\
        \rowcolor{Gray}
        TCFormerV2-Small & $25.6$ & $4.5$ & $237.5$ & $82.4$ \\
	\hline
	ResNet101~\cite{he2016deep}  & $44.7$ & $7.9$ & $689.0$& $77.4$\\
	ResNeXt101-32x4d~\cite{xie2017aggregated} & $44.2$ & $8.0$ & $495.9$ & $78.8$ \\
	RegNetY-8G~\cite{radosavovic2020designing} & $39.0$ & $8.0$ & $564.3$ & $81.7$ \\
	T2T-ViT$_t$-19~\cite{yuan2021tokens} & $39.0$ & $9.8$ & $-$ & $81.4$ \\
	PVTv1-Medium~\cite{wang2021pyramid} & $44.2$ & $6.7$ & $516.1$ & $81.2$\\
	CvT-21~\cite{wu2021cvt} & $32.0$ & $7.1$ & $-$ & $82.5$ \\
        iFormer-S~\cite{si2022inception} & $48.0$ & $9.4$ & $-$ & $84.6$ \\
	\hline
	ResNet152~\cite{he2016deep} & $60.2$ & $11.6$ & $463.7$ & $78.3$ \\
	T2T-ViT$_t$-24~\cite{yuan2021tokens} & $64.0$ & $15.0$ & $-$ & $82.2$ \\
	PVTv1-Large~\cite{wang2021pyramid} & $61.4$ & $9.8$ & $348.3$ & $81.7$ \\
	TNT-B~\cite{han2021transformer} & $66.0$ & $14.1$ & $246.7$ & $82.8$ \\
	Swin-S~\cite{liu2021swin} & $50.0$ & $8.7$ & $431.7$ & $83.0$ \\
	Twins-SVT-B~\cite{chu2021twins} & $56.0$ & $8.3$ & $466.7$ & $83.2$ \\
	PVTv2-B4~\cite{wang2022pvt} & $62.6$ & $10.1$ & $320.4$ & $83.6$ \\
        \rowcolor{Gray}
        TCFormerV1-Large  & $62.8$ & $12.2$ & $58.7$ & $83.6$ \\
        \rowcolor{Gray}
        TCFormerV2-Base  & $62.8$ & $10.8$ & $103.0$ & $83.8$ \\
        \Thline
    \end{tabular}
    }
\end{table}

\subsection{Image Classification} \label{sec:cls}

\textbf{Settings.}
%
We train our TCFormer on the ImageNet-1K dataset~\cite{russakovsky2015imagenet}, which comprises 1.28 million training images and 50,000 validation images across 1,000 categories. 
The experimental settings are consistent with PVT~\cite{wang2021pyramid}. We employ data augmentations of random cropping, random horizontal flipping~\cite{szegedy2015going}, label-smoothing~\cite{szegedy2016rethinking}, Mixup~\cite{zhang2017mixup}, CutMix~\cite{yun2019cutmix}, and random erasing~\cite{zhong2020random}.  All models are trained from scratch for 300 epochs with a batch size of 128. The models are optimized with an AdamW~\cite{loshchilov2018decoupled} optimizer with a momentum of $0.9$ and a weight decay of $5 \times 10^{-2}$. The initial learning rate is set to $1 \times 10^{-3}$  and decreases following the cosine schedule~\cite{loshchilov2016sgdr}. We evaluate our models on the validation set with a center crop of $224 \times 224$ patch.

\begin{figure*}[tb]
	\centering
        \includegraphics[width=0.96\textwidth]{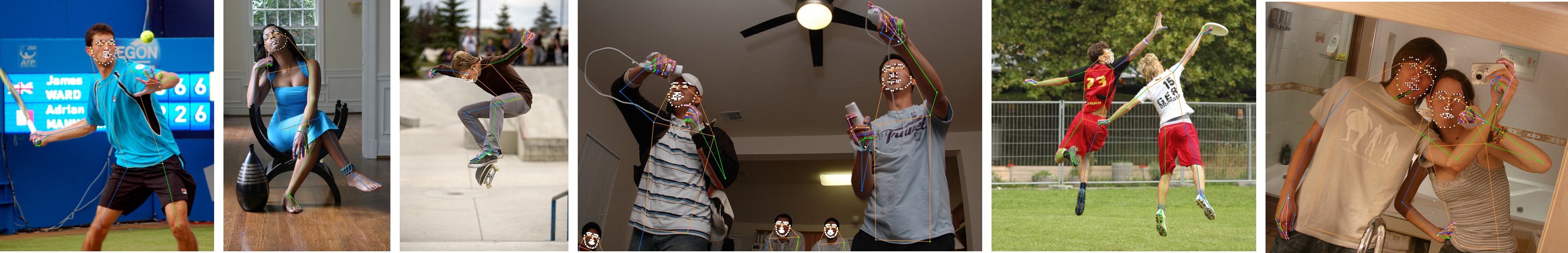}
        \caption{
        Qualitative results of whole-body pose estimation (heatmap-based) on COCO-WholeBody dataset.
        }
	\label{fig:wholebody_vis}
\end{figure*}

\textbf{Results.}
We present a comparison of our proposed TCFormer with state-of-the-art approaches on the ImageNet-1K \texttt{val} set in Table~\ref{tab:cls}.
TCFormerV1 exhibits superiority over traditional convolutional backbones and transformer backbones.  Specifically, TCFormerV1 attains a top-1 accuracy of $82.4\%$, which is $6.3$ point higher than ResNet50~\cite{he2016deep}, $1.1\%$ higher higher than Swin-T~\cite{liu2021swin} and $2.6\%$ higher than PVT~\cite{wang2021pyramid}. It proves the effectiveness of our dynamic tokens.
However, due to the global clustering process, despite having a parameter number comparable to other state-of-the-art models, TCFormerV1 exhibits relatively high computational complexity.
Conversely, TCFormerV2 delivers similar performance to TCFormerV1 but with a significantly lower computational complexity. For instance, TCFormerV2-Small achieves the same performance as TCFormerV1 but has $23.7\%$ fewer GFLOPs. Similar trends are observed across other model scales, highlighting the efficiency and effectiveness of our newly proposed local CTM module in TCFormerV2.

In comparison with methods integrating enhanced transformer blocks, such as iFormer~\cite{si2022inception}, TCFormer achieves comparable performance. 
The combination of our dynamic tokens with more advanced transformer blocks may present a promising avenue for achieving superior performance.
Despite the reduction in computational complexity achieved by the local CTM module, TCFormer still exhibits a lower throughput than methods with fixed grid structures. 
This limitation arises from the inherent incompatibility of our dynamic tokens with existing grid-based convolutional layers. The convolution layers employed in TCFormer results in the time-consuming transformation between dynamic tokens and feature maps. Addressing this challenge requires the development of new transformer modules friendly to dynamic tokens or the introduction of new hardware support.

\begin{table*}[t]
	\caption{ OKS-based Average Precision (AP) and Average Recall (AR) on the COCO-WholeBody V1.0 dataset. The baseline results are from MMPose~\cite{mmpose2020}. `*' indicates multi-scale testing. ZoomNet$^{\dagger}$ is trained with the COCO-WholeBody V0.5 training set.
	}
	\label{tab:wholebody}
	\begin{center}
    \scalebox{0.94}{
		\begin{tabular}{c|c|c|cc|cc|cc|cc|aa}
		\hline
			 \multirow{2}{*}{Method} & \multirow{2}{*}{Resolution} & \multirow{2}{*}{GFLOPs} &  \multicolumn{2}{c|}{body}  & \multicolumn{2}{c|}{foot}  & \multicolumn{2}{c|}{face}  & \multicolumn{2}{c|}{hand} & \multicolumn{2}{c}{\cellcolor{Gray}whole-body} \\
			\cline{4-13}
		  ~ & ~ & ~ & AP     & AR     & AP   & AR     &  AP  & AR     & AP    & AR   &  AP     & AR  \\
			\hline
            SN$^{*}$~\cite{hidalgo2019single} & $480\times480$ & 272.3 & 0.427 & 0.583 & 0.099 & 0.369 & 0.649 & 0.697 & 0.408 & 0.580 & 0.327 & 0.456 \\ 
            OpenPose~\cite{cao2018openpose} & $480\times480$ & 451.1 & 0.563 & 0.612 & 0.532 & 0.645 & 0.765 & 0.840 & 0.386 & 0.433 & 0.442 & 0.523 \\ 
            PAF$^{*}$~\cite{cao2017realtime} & $480\times480$ & 329.1 & 0.381 & 0.526 & 0.053 & 0.278 & 0.655 & 0.701 & 0.359 & 0.528 & 0.295 & 0.405 \\ 
            AE~\cite{newell2017associative}+HRNet-w48~\cite{sun2019deep} & $512\times512$ & 212.4 & 0.592	& 0.686& 	0.443& 	0.595& 	0.619& 	0.674& 	0.347& 	0.438& 	0.422& 	0.532 \\ 
            HigherHRNet-w48~\cite{cheng2020higherhrnet} & $512\times512$ & 99.7 & 0.630 &	0.706 &	0.440 &	0.573 &	0.730 &	0.777 &	0.389 &	0.477 &	0.487 &	0.574 \\ 
            \hline
            SBL-Res50~\cite{xiao2018simple}  & $256\times192$ & 5.6 & 0.652 &	0.739 &	0.614 &	0.746 &	0.608 &	0.716 &	0.460 &	0.584 &	0.520 &	0.633 \\
            HRNet-w32~\cite{sun2019deep} & $256\times192$ & 7.7 & 0.700 &	0.746 &	0.567 &	0.645 &	0.637 &	0.688 &	0.473 &	0.546 &	0.553 &	0.626 \\ 
            TCFormerV1~\cite{zeng2022not} & $256\times192$ & 8.7 & 0.691 & 0.770 & 0.698 & 0.813 & 0.649 & 0.746 & 0.535 & 0.650 & 0.572 & 0.678 \\ 
            TCFormerV2-Small & $256\times192$ & 7.3 & 0.695 & 0.773 & 0.694 & 0.813 & 0.652 & 0.748 & 0.535 & 0.649 & \textbf{0.575} & \textbf{0.682} \\ 
            \hline
            ZoomNet$^{\dagger}$~\cite{jin2020whole} & $384\times288$ & 27.4 & 0.743 & 0.802 &  0.798 & 0.869 & 0.623 & 0.701 & 0.401 & 0.498 & 0.541 & 0.658 \\ 
            SBL-Res152~\cite{xiao2018simple}   & $384\times288$ & 28.9 & 0.703	&0.780	&0.693&	0.813&	0.751&	0.825&	0.559&	0.667&	0.610&	0.705 \\
            HRNet-w48~\cite{sun2019deep} & $384\times288$ & 35.5 & 0.722 &	0.790&	0.694&	0.799&	0.777	&0.834&	0.587	&0.679	&0.631	&0.716 \\ 
            TCFormerV1-Large~\cite{zeng2022not} & $384\times288$ &38.7 & 0.731 & 0.803 & 0.752 & 0.855 & 0.774 & 0.845 & 0.607 & 0.712 & 0.644 & 0.735 \\ 
            TCFormerV2-Base & $384\times288$ & 31.7 & 0.727 & 0.804 & 0.746 & 0.856 & 0.788 & 0.858 & 0.626 & 0.728 & \textbf{0.651} & \textbf{0.742} \\ 
		\hline
		\end{tabular}
		}
	\end{center}
\end{table*}

\subsection{Human Pose Estimation} \label{sec:pose}
%
Human pose estimation aims to localize predefined keypoints, also known as body joints, in the input images. Current approaches can be divided into two categories: heatmap-based methods and regression-based methods. To thoroughly assess the performance of TCFormer, we apply TCFormer backbones to both heatmap-based and regression-based algorithms.

\subsubsection{Heatmap Based Method}
\textbf{Settings.}
We conduct experiments on the COCO-WholeBody V1.0 dataset~\cite{jin2020whole,xu2022zoomnas}. COCO-WholeBody dataset is a large-scale 2D whole-body pose estimation benchmark built upon the well-known COCO dataset~\cite{lin2014microsoft} and contains over 200K instance annotations for 133 predefined keypoints, including 17 for the body, 6 for the feet, 68 for the face, and 42 for the hands. Following~\cite{lin2014microsoft,jin2020whole,xu2022zoomnas}, we evaluate the model performance using OKS-based Average Precision (AP) and Average Recall (AR). We adopt the default training and evaluation settings of MMPose~\cite{mmpose2020} and only replace the Adam optimizer~\cite{kingma2014adam} with an AdawW optimizer~\cite{loshchilov2018decoupled} with a momentum of $0.9$ and a weight decay of $1 \times 10^{-2}$.

\begin{table*}[t]
\centering
\caption{Human pose estimation on MS-COCO \texttt{val} set.}
\label{tab:coco_pose_val}
\begin{tabular}{l|c|c|c|c|c|c|c|c|c|c}
\toprule
\renewcommand{\arraystretch}{0.1}
Method & Backbone & Input Size & \#Param & GFLOPs &  AP   & $\text{AP}^{50}$ & $\text{AP}^{75}$  & $\text{AP}^{M}$ & $\text{AP}^{L}$  &AR \\
\midrule
\multicolumn{2}{l}{\textit{Heatmap-based}} \\
Hourglass~\cite{newell2016stacked} & $8$-stage Hourglass &   $256 \times 192$ & $25.1$M & $14.3$&
$66.9$&$-$&$-$&$-$&$-$&$-$\\ 
CPN~\cite{chen2018cascaded}& ResNet50 &  $256 \times 192$ & $27.0$M & $6.2$&
$68.6$&$-$&$-$&$-$&$-$&$-$\\ 
SimpleBaseline~\cite{xiao2018simple} & ResNet50 & $256\times192$  &$34.0$M &$8.9$
&${70.4}$ & ${88.6}$&${78.3}$&${67.1}$&${77.2}$&${76.3}$\\
SimpleBaseline~\cite{xiao2018simple} & ResNet152  & $256\times192$ &$68.6$M &$15.7$
&${72.0}$ & ${89.3}$&${79.8}$&${68.7}$&${78.9}$&${77.8}$\\
HRNet~\cite{sun2019deep} & HRNet-W$32$&  $256\times 192$&  $28.5$M & $7.1$ &
$74.4$&$90.5$&$81.9$&$70.8$&$81.0$&$79.8$  \\ 
HRNet~\cite{sun2019deep} & HRNet-W$48$&  $256\times 192$&  $63.6$M &$14.6$ &
$75.1$&$90.6$&$82.2$&$71.5$&$81.8$&$80.4$  \\
SimpleBaseline~\cite{xiao2018simple} & ResNet152  &   $384\times288$     &$68.6$M &$35.6$
&${74.3}$ & ${89.6}$&${81.1}$&${70.5}$&${79.7}$&${79.7}$\\
HRNet~\cite{sun2019deep} & HRNet-W$48$&   $384\times 288$&  $63.6$M &$32.9$ & $76.3$&$90.8$&$82.9$&$72.3$&$83.4$&$81.2$  \\
\midrule
\multicolumn{2}{l}{\textit{Regression-based}} \\
DeepPose~\cite{toshev2014deeppose} & ResNet50 & $256 \times 192$ & $-$ & $-$ & $52.6$&	$81.6$&	$58.6$& $50.0$&	$59.1$&	$63.8$  \\
TransPose-R-A3~\cite{yang2021transpose} & ResNet50 & $256 \times 192$ & $-$ & $8.0$ & $71.7$ &	$88.9$&	$78.8$&	$68.0$&	$78.6$&	$77.1$ \\
TransPose-R-A4~\cite{yang2021transpose} & ResNet50 & $256 \times 192$ & $-$ & $8.9$ & $72.6$	&$89.1$&	$79.9$&	$68.8$&	$79.8$&	$78.0$  \\
RLE~\cite{li2021human}+ResNet~\cite{he2016deep} & ResNet50 & $256 \times 192$ & $23.7$M & $4.0$ & $70.4$	& $88.3$&	$77.7$ & $70.9$ & $81.1$ & 	$75.1$    \\
\rowcolor{Gray}
RLE~\cite{li2021human}+TCFormer & TCFormerV2-Small & $256 \times 192$ & $25.2$M & $4.4$ & $73.9$ & $89.9$ & $81.0$ & $70.4$ & $80.4$ & $78.6$\\
RLE~\cite{li2021human}+ResNet~\cite{he2016deep} & ResNet152 & $256 \times 192$ & $58.3$M & $11.3$ &  $73.1$	&$89.7$&	$80.5$&	$73.7$ & $83.6$ & $77.7$   \\
\rowcolor{Gray}
RLE~\cite{li2021human}+TCFormer & TCFormerV2-Base & $256 \times 192$  & $62.4$M & $10.6$ & $76.0$ & $90.7$ & $82.9$ & $72.4$ & 82.3 & $80.5$  \\
RLE~\cite{li2021human}+ResNet~\cite{he2016deep} & ResNet152 & $384 \times 288$ & $58.3$M & $25.5$ &  $74.9$ & $90.1$  &$81.5$  & $75.0$ & $85.5$ & $79.3$   \\
\rowcolor{Gray}
RLE~\cite{li2021human}+TCFormer & TCFormerV2-Base & $384 \times 288$ & $62.4$M & $25.1$ & $77.1$ & $91.0$ & $83.7$ & $73.4$ & $83.7$ & $81.5$\\
\bottomrule
\end{tabular}
\end{table*}

\textbf{Results.}
Table~\ref{tab:wholebody} presents the results on COCO-WholeBody V1.0 dataset~\cite{xu2022zoomnas}. We compare TCFormer with previous state-of-the-art methods, such as HRNet~\cite{sun2019deep} and ZoomNet~\cite{jin2020whole}. TCFormerV1 surpasses previous state-of-the-art methods by a large margin, while TCFormerV2 further improves the performance and reduces the computational complexity. With the input resolution of $256\times192$, TCFormerV2-Small achieves the performance of $57.5\%$ AP and $68.2\%$ AR, which is $2.2\%$ AP and $5.6\%$ AR higher than HRNet-w32. 
With a higher input resolution and larger model, TCFormerV2-Base achieves new state-of-the-art performance of $65.1\%$ AP and $74.2\%$ AR, surpassing HRNet48 by $2.0\%$ AP and SBL-Res152~\cite{xiao2018simple} by $4.1\%$ AP. 
The improvement in TCFormer is attributed to its superior detail capture ability. Human hands have a complex structure but typically occupy a small area in the input images, presenting a challenge for models to reconstruct hand keypoints. 
As seen in Table~\ref{tab:wholebody}, most models perform much worse on hand keypoint estimation than other body parts.
In contrast, our TCFormer can capture image details better, as it represents details with finer vision tokens, resulting in a significant improvement in hand keypoint estimation.
Specifically, TCFormerV2-Small outperforms HRNet-w32 by $6.2\%$ AP, and TCFormerV2-Base outperforms HRNet-w48 by $3.9\%$ AP on hand keypoints.
We show some qualitative results of TCFormer in Fig.~\ref{fig:wholebody_vis} showcases some qualitative results of TCFormer-Base.

\begin{figure*}[tb]
	\centering
	\includegraphics[width=0.96\textwidth]{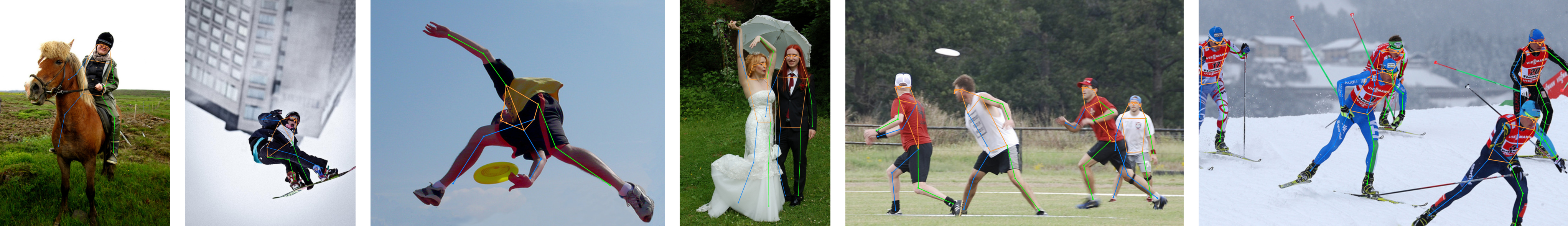}
        \caption{
        Qualitative results of pose estimation (regression-based) on COCO dataset.
        }
	\label{fig:pose_vis}
\end{figure*}

\begin{table*}[t]
    \caption{Human pose estimation on MS-COCO \texttt{test-dev} set.}
\centering
\label{table:coco_test_dev}
\footnotesize
\begin{tabular}{l|l|c|c|c|lllllc}
    \hline
    Method &Backbone& Input size & \#Params & GFLOPs&
    $\operatorname{AP}$ & $\operatorname{AP}^{50}$ & $\operatorname{AP}^{75}$ & $\operatorname{AP}^{M}$ & $\operatorname{AP}^{L}$ & $\operatorname{AR}$\\
    \hline
    \multicolumn{2}{l}{\textit{Heatmap-based}} \\
    OpenPose~\cite{cao2018openpose} & 3CM-3PAF (102) & $-$ &$-$& $-$  
    &$61.8$ & $84.9$&$67.5$&$57.1$&$68.2$&$66.5$\\
    Associative Embedding~\cite{newell2017associative} & 4-stage Hourglass  & $-$ &$-$& $-$
    &$65.5$ & $86.8$&$72.3$&$60.6$&$72.6$&$70.2$\\
    HGG~\cite{jin2020differentiable} & 4-stage Hourglass & $-$ & $-$ &$-$ & $67.6$ & $85.1$ &  $73.7$ & $62.7$ &  $74.6$ & $71.3$ \\
    Mask-RCNN~\cite{he2017mask} & ResNet-50-FPN& $-$ &$-$& $-$
    & $63.1$ & $87.3$&$68.7$&$57.8$&$71.4$&$-$\\
    G-RMI~\cite{papandreou2017towards} & ResNet-101 & $353\times257$ &$42.6$M& $57.0$
    &$64.9$ & $85.5$&$71.3$&$62.3$&$70.0$&$69.7$\\
    Integral Pose~\cite{sun2018integral} & ResNet-101 & $256\times256$ &$45.0$M& $11.0$
    &$67.8$ & $88.2$&$74.8$&$63.9$&$74.0$&$-$\\
    CPN~\cite{chen2018cascaded} & ResNet-Inception& $384\times288$ &$58.8$M& $29.2$
    & $72.1$ & $91.4$&$80.0$&$68.7$&$77.2$&$78.5$\\
    SimpleBaseline~\cite{xiao2018simple} & ResNet-152&$384\times288$  &$68.6$M& $35.6$
    &${73.7}$ & ${91.9}$&${81.1}$&${70.3}$&${80.0}$&${79.0}$\\
    TransPose~\cite{yang2021transpose} &  TransPose-H-A$6$ & $256\times 192$          & $17.5$M                  & $21.8$                 & ${75.0}$                & ${92.2}$            & ${82.3}$ & ${71.3}$ & ${81.1}$ & ${-}$    \\
    HRNet~\cite{sun2019deep}  &   HRNet-W$48$                 & $384\times 288$          & $63.6$M                  & $32.9$                 & ${75.5}$                & ${92.5}$            & ${83.3}$ & ${71.9}$ & ${81.5}$ & ${80.5}$ \\
    TokenPose~\cite{li2021tokenpose} &   TokenPose-L/D$24$  & $384\times 288$          & $29.8$M                  & $22.1$                 & ${75.9}$                & ${92.3}$            & ${83.4}$ & ${72.2}$ & ${82.1}$ & ${80.8}$ \\
    \hline
    \multicolumn{2}{l}{\textit{Regression-based}} \\
    CenterNet~\cite{zhou2019objects} & Hourglass-104 &$-$&$-$&$-$& $63.0$ & $86.8$ & $69.6$ & $58.9$ & $70.4$ & $-$ \\
    PointSet Anchor~\cite{wei2020point} & HRNet-W48 &$-$&$-$&$-$& $68.7$ & $89.9$ & $76.3$ & $64.8$ & $75.3$ & $-$\\
    PRTR~\cite{li2021pose}   &   PRTR  & $512\times 384$   & $57.2$M   & $37.8$ & ${72.1}$   & ${90.4}$   & ${79.6}$ & ${68.1}$ & ${79.0}$ & ${79.4}$ \\
    RLE~\cite{li2021human}+ResNet~\cite{he2016deep} & ResNet-152 & $384\times 288$  & $58.3$M & $25.5$ & $74.2$ & $91.5$ & $81.9$ & $71.2$ & $79.3$ & $-$  \\
    RLE~\cite{li2021human}+HRNet~\cite{sun2019deep} & HRNet-W48 & $384\times 288$  & $75.6$M & $38.3$ & $75.7$ & $92.3$ & $82.9$ & $72.3$ & $81.3$ & $-$ \\
    \rowcolor{Gray}
    RLE~\cite{li2021human}+TCFormer & TCFormerV2-Base & $384\times 288$  & $62.4$M & $25.1$ & $76.1$ & $92.4$ & $83.7$ & $72.7$ & $81.8$ & $86.1$ \\
    \hline
\end{tabular}
\end{table*}

\subsubsection{Regression Based Method}
\textbf{Settings.} 
We choose COCO~\cite{lin2014microsoft} dataset to evaluate the performance of TCFormer on regression-based human pose estimation. COCO is a large-scale human pose estimation dataset with over 250K instance annotations for 17 predefined human keypoints. We apply TCFormer backbones to the RLE~\cite{li2021human} framework and follow the default training and evaluation settings of MMPose~\cite{mmpose2020}. All compared methods use the weights pre-trained on ImageNet-1K~\cite{russakovsky2015imagenet} dataset. Since RLE is a regression-based method, it does not require dense output. Hence, we do not use the MTA module in the experiments. Instead, we directly regress the keypoint locations from the mean token features in the last stage, which is similar to the setting of image classification.

\textbf{Results.}  
We present some qualitative results of TCFormer in Fig.~\ref{fig:pose_vis}.
Table~\ref{tab:coco_pose_val} shows the comparisons between TCFormer and other state-of-the-art methods on the COCO val set. TCFormer outperforms both convolution-based models (RLE~\cite{li2021human} + ResNet~\cite{he2016deep}) and transformer-based models (TransPose~\cite{yang2021transpose}) by a large margin. With similar computational complexity, RLE + TCFormerV2-Base outperforms RLE + ResNet152 by $2.9\%$ AP and outperforms TransPose-R-A4 by $3.4\%$ AP. With higher input resolution, RLE + TCFormerV2-Base achieves a new state-of-the-art performance of $77.1\%$ AP.
We also report the performance of TCFormer on the COCO test set in Table~\ref{table:coco_test_dev}. RLE + TCFormerV2-Base outperforms other state-of-the-art regression-based methods with $76.1\%$ AP.
It is worth noting that while RLE + HRNet uses a dense feature map and a more sophisticated prediction head, RLE + TCFormer only uses a simple regression head.

\begin{table}[t]
\centering
\caption{\textbf{Comparisons with PVTs on semantic segmentation on the ADE20K validation set.}
    ``GFLOPs'' is calculated under the input scale of $512\times 512$.
}
\label{tab:seg_fpn}
\begin{tabular}{l|c|c|c}
\Thline
\renewcommand{\arraystretch}{0.1}
\multirow{2}{*}{Backbone} & \multicolumn{3}{c}{Semantic FPN}\\
\cline{2-4}
& \#Param & GFLOPs & mIoU (\%)   \\


\hline
ResNet50~\cite{he2016deep} & $28.5$M & $45.6$ & $36.7$\\
PVTv1-Small~\cite{wang2021pyramid} & $28.2$M & $44.5$ & $39.8$ \\
PVTv2-B2~\cite{wang2022pvt} & $29.1$M & $45.8$ & $45.2$ \\
\rowcolor{Gray}
TCFormerV1 & $29.4$M & $92.4$ & $47.1$ \\
\rowcolor{Gray}
TCFormerV2-Small & $28.4$M & $44.4$ & $47.8$ \\


\hline
ResNeXt101-64x4d~\cite{xie2017aggregated} &$86.4$M & $103.9$ & 40.2 \\
PVTv1-Large~\cite{wang2021pyramid} & $65.1$M & $79.6$ & $42.1$ \\
PVTv2-B4~\cite{wang2022pvt} & $66.3$M & $81.3$ & $47.9$ \\
\rowcolor{Gray}
TCFormerV2-Base & $66.0$M & $84.0$ & $50.0$ \\


\Thline
\end{tabular}

\end{table}

\begin{table}[t]
\centering
\setlength{\tabcolsep}{1.5mm}
\caption{\textbf{Comparisons with Swin Transformers on semantic segmentation on the ADE20K validation set.}
}
\label{tab:seg_mask2former}
\begin{tabular}{l|c|c|c|c}
\Thline
\renewcommand{\arraystretch}{0.1}
\multirow{2}{*}{Backbone} & \multicolumn{4}{c}{Mask2Former}\\
\cline{2-5}
&   Input size & \#Param & GFLOPs & mIoU (\%)   \\
\Thline
ResNet50~\cite{he2016deep} & $512\times512$ & $44.0M$& $70.8$ & $47.2$ \\
Swin-T~\cite{liu2021swin} & $512\times 512$ & $47.4$M & $73.6$ & $47.7$\\
\rowcolor{Gray}
TCFormerV2-Small & $512\times 512$ & $42.4$M & $56.8$ & $49.1$ \\
\hline
ResNet101~\cite{he2016deep} & $512\times 512$ & $63.0M$& $90.2$ & $47.8$\\
Swin-S~\cite{liu2021swin} & $512\times 512$ &  $68.8$M & $97.4$  & $51.3$ \\
\rowcolor{Gray}
TCFormerV2-Base & $512\times 512$ & $79.5$M & $93.7$ & $52.8$ \\
\hline
Swin-B~\cite{liu2021swin} & $640\times 640$ & $86.9$M &  $223.3$ & $52.4$ \\
\rowcolor{Gray}
TCFormerV2-Base & $640\times 640$ &  $79.5$M & $155.4$ & $53.8$ \\
\Thline
\end{tabular}

\end{table}

\begin{figure*}[tb]
	\centering
	\includegraphics[width=0.96\textwidth]{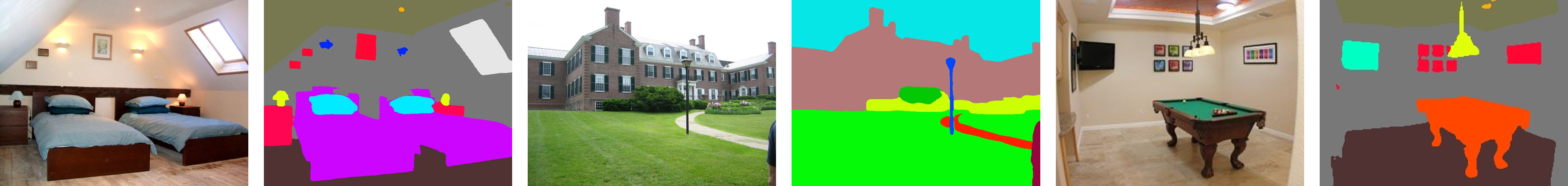}
        \caption{
        Qualitative results of semantic segmentation (TCFormer + Semantic FPN) on ADE20K dataset.
        }
	\label{fig:seg_vis}
\end{figure*}

\subsection{Semantic Segmentation} \label{sec:seg}

\textbf{Settings.}
We perform our experiments on the ADE20K dataset~\cite{zhou2017scene}, a large-scale benchmark for semantic segmentation. ADE20K consists of 25K images annotated with 150 categories, and is split into 20K, 2K, and 3K images for training, validation, and testing, respectively.
Our approach, TCFormer, is applied to two different frameworks: the classic convolution-based framework Semantic FPN~\cite{kirillov2019panoptic}, and the more recent transformer-based framework Mask2Former~\cite{cheng2022masked}. In all cases, we utilize backbones initialized with weights pre-trained on ImageNet-1K
For Semantic FPN framework, we use TCFormer as the backbone and replace the FPN~\cite{lin2017feature} module with our proposed MTA module. We follow the experimental settings of PVT~\cite{wang2022pvt}. During training, images are are randomly resized and cropped to $512 \times 512$. While in evaluation, images are resized such that the shorter side has $512$ pixels. We train our models for 40k iterations with a batch size of 16 and utilize an AdamW optimizer for model optimization. The initial learning rate is set to $1 \times 10^{-4}$ and decays following a polynomial decay schedule with a power of $0.9$. 
For Mask2Former framework, we implement TCFormer as the backbone and replace the pixel decoder with our MTA module. We follow the experiment setting of \cite{cheng2022masked}. All models are optimized for 160k iterations using an AdamW optimizer with an initial learning rate of $1 \times 10^(-4)$ and a weight decay of $0.05$. We utilize the poly schedule to decay the learning rate.

\textbf{Results.}
Qualitative results of TCFormer with Semantic FPN framework are shown in Fig.~\ref{fig:seg_vis}.
%
%
In Table~\ref{tab:seg_fpn}, we present a comparison of TCFormer with other state-of-the-art methods based on the Semantic FPN framework. Our approach, TCFormer, outperforms both CNN models (ResNet~\cite{he2016deep}) and transformer models (PVT~\cite{wang2021pyramid}) by a significant margin. Specifically, the mIoU of TCFormerV2-Small is $11.1$ points higher than ResNet50 and $2.6$ points higher than PVTv2-B2~\cite{wang2022pvt}. However, the global token clustering approach used in TCFormerV1 results in unacceptable computational complexity when the input resolution is $512 \times 512$, as reflected in the huge GFLOPs of TCFormerV1 models in Table~\ref{tab:seg_fpn}. With the use of the local CTM modules, TCFormerV2 achieves better performance while saving on large computational complexity in all model scales. 
Compared to TCFormerV1, TCFormerV2-Small achieves a performance gain of $0.7\%$ mIoU while using only $48.1\%$ GFLOPs.

%
In Table~\ref{tab:seg_mask2former}, we show the results of TCFormer based on the Mask2Former framework. TCFormer outperforms ResNet and Swin models by a large margin in all model scales. The dynamic vision tokens used in TCFormer enable the model to allocate computation cost according to image semantic meaning, making TCFormer more efficient than traditional vision transformers. Specifically, with the resolution of $640 \times 640$, TCFormerV2-Base outperforms Swin-B by $1.4\%$ mIoU while using $30.4\%$ less GFLOPs.

\begin{figure*}[tb]
	\centering
	\includegraphics[width=0.96\textwidth]{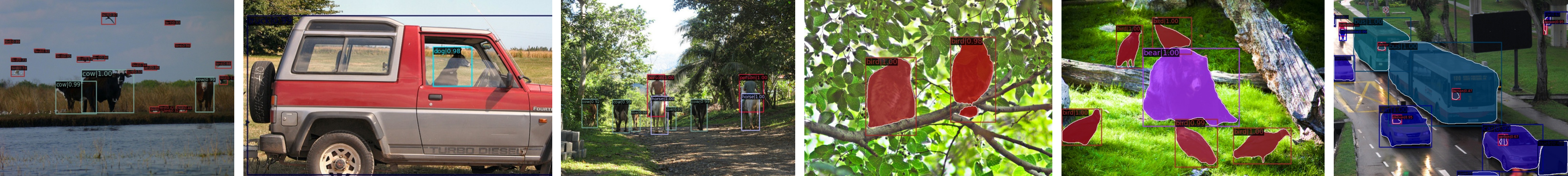}
        \caption{
        Qualitative results of object detection (left) and instance segmentation (right) on COCO dataset.
        }
	\label{fig:det_vis}
\end{figure*}

\begin{table}[t]
\scriptsize
\centering
\renewcommand{\arraystretch}{1}
\setlength{\tabcolsep}{1.3mm}
\caption{Object detection and instance segmentation results on COCO val2017. TCFormer is compared with other backbones on RetinaNet and Mask R-CNN frameworks.}
\label{table:det}
\begin{tabular}{l|c|ccc|ccc}
    \toprule[1.5pt]
    \multirow{2}{*}{Backbone} & \multicolumn{7}{c}{RetinaNet 1$\times$}  \\
    \cline{2-8}
    & \#Param &  mAP & AP$_{50}$ &AP$_{75}$  &AP$_{\rm S}$ &AP$_{\rm M}$ & AP$_{\rm L}$\\
    
    \midrule	
    ResNet18~\cite{he2016deep} & 21.3M &  31.8 & 49.6 & 33.6 & 16.3 & 34.3 & 43.2 \\
    PVT-Tiny~\cite{wang2021pyramid} & 23.0M & 36.7  & 56.9 & 38.9 & 22.6 & 38.8 & 50.0 \\
    PVTv2-B1~\cite{wang2022pvt} & 23.8M  & 41.2 & 61.9 & 43.9 & 25.4 & 44.5 & 54.3 \\
    \rowcolor{Gray}
    TCFormerV2-Tiny & 23.1M & 41.7 & 62.4 & 44.3 & 25.4 & 45.1 & 54.4 \\
    
    \midrule		
    ResNet50~\cite{he2016deep} & 37.7M &  36.3 & 55.3 & 38.6 & 19.3 & 40.0 & 48.8 \\
    PVT-Small~\cite{wang2021pyramid} & 34.2M  & 40.4 & 61.3 & 44.2 & 25.0 & 42.9 & 55.7 \\
    PVTv2-B2~\cite{wang2022pvt} & 35.1M & 44.6 & 65.6 & 47.6 & 27.4 & 48.8 & 58.6 \\
    CycleMLP-B2~\cite{chen2021cyclemlp} & 36.5M & 40.6 & 61.4 & 43.2 & 22.9 & 44.4 & 54.5 \\
    Swin-T~\cite{liu2021swin} & 38.5M & 41.5 & 62.1 & 44.2 & 25.1 & 44.9 & 55.5 \\
    Pyramid ViG-S~\cite{han2022vision} & 36.2M & 41.8 & 63.1 & 44.7 & 28.5 & 45.4 & 53.4 \\
    \rowcolor{Gray}
    TCFormerV2-Small & 34.5M & 45.0 & 66.2 & 48.1 & 28.8 & 48.2 & 60.0 \\

    \midrule
    PVT-Large~\cite{wang2021pyramid} & 71.1M  & 42.6 & 63.7 & 45.4 & 25.8 & 46.0 & 58.4 \\
    PVTv2-B4~\cite{wang2022pvt} & 72.3M & 46.1 & 66.9 & 49.2 & 28.4 & 50.0 & 62.2 \\
    \rowcolor{Gray}
    TCFormerV2-Base & 72.4M & 46.6 & 67.6 & 50.0 & 29.7 & 50.5 & 61.9 \\
    \midrule
    
    \multirow{2}{*}{Backbone}  &\multicolumn{7}{c}{Mask R-CNN 1$\times$} \\
    \cline{2-8}
    & \#Param & AP$^{\rm b}$ & AP$_{50}^{\rm b}$ &AP$_{75}^{\rm b}$  &AP$^{\rm m}$ &AP$_{50}^{\rm m}$ & AP$_{75}^{\rm m}$\\
    
    \midrule		
    ResNet18~\cite{he2016deep} &  31.2M & 34.0 & 54.0 & 36.7 & 31.2 & 51.0 & 32.7\\
    PVT-Tiny~\cite{wang2021pyramid} & 32.9M & 36.7 & 59.2 & 39.3 & 35.1 & 56.7 & 37.3\\
    PVTv2-B1~\cite{wang2021pyramid} & 33.7M & 41.8 & 64.3 & 45.9 & 38.8 & 61.2 & 41.6\\
    \rowcolor{Gray}
    TCFormerV2-Tiny & 34.1M & 42.6 & 65.1 & 46.8 & 39.2 & 62.1 & 42.0  \\
    
    \midrule		
    ResNet50~\cite{he2016deep} & 44.2M & 38.0 & 58.6 & 41.4 & 34.4 & 55.1 & 36.7 \\
    PVT-Small~\cite{wang2021pyramid} & 44.1M & 40.4 & 62.9 & 43.8 & 37.8 & 60.1 & 40.3 \\
    PVTv2-B2~\cite{wang2022pvt} &  45.0M & 45.3 & 67.1 & 49.6 & 41.2 & 64.2 & 44.4\\
    CycleMLP-B2~\cite{chen2021cyclemlp} & 46.5M & 42.1 & 64.0 & 45.7 & 38.9 & 61.2 & 41.8 \\
    PoolFormer-S24~\cite{yu2022metaformer} & 41.0M  & 40.1 & 62.2 & 43.4 & 37.0 & 59.1 & 39.6 \\
    Swin-T~\cite{liu2021swin} & 47.8M & 42.2 & 64.6 & 46.2 & 39.1 & 61.6 & 42.0 \\
    Pyramid ViG-S~\cite{han2022vision} & 45.8M  & 42.6 & 65.2 & 46.0 & 39.4 & 62.4 & 41.6 \\
    \rowcolor{Gray}
    TCFormerV2-Small & 46.5M & 46.1 & 68.4 & 50.5 & 41.9 & 65.4 & 45.1 \\
    
    \midrule	
    \multirow{2}{*}{Backbone}  &\multicolumn{7}{c}{Mask R-CNN 3$\times$} \\
    \cline{2-8}
    & \#Param & AP$^{\rm b}$ & AP$_{50}^{\rm b}$ &AP$_{75}^{\rm b}$  &AP$^{\rm m}$ &AP$_{50}^{\rm m}$ & AP$_{75}^{\rm m}$\\

    \midrule
    PVT-Small~\cite{wang2021pyramid} & 44.1M & 43.0 & 65.3 & 46.9 & 39.9 & 62.5 & 42.8 \\
    Swin-T~\cite{liu2021swin} & 47.8M & 46.0 & 68.1 & 50.3 & 41.6 & 65.1 & 44.9 \\
    \rowcolor{Gray}
    TCFormerV2-Small & 46.5M & 48.2 & 70.1 & 53.4 & 43.4 & 67.2 & 46.9  \\

    \midrule
    PVT-Large~\cite{wang2021pyramid} & 81.0M & 44.5 & 66.0 & 48.3 & 40.7 & 63.4 & 43.7 \\
    Swin-S~\cite{liu2021swin} & 69.0M & 48.5 & 70.2 & 53.5 & 43.3 & 67.3 & 46.6 \\
    \rowcolor{Gray}
    TCFormerV2-Base & 83.7M & 49.8 & 71.0 & 54.7 & 44.4 & 68.2 & 48.0 \\
    \bottomrule[1pt]
\end{tabular}
\end{table}

\subsection{Object Detection} \label{sec:det}

\textbf{Settings.} 
%
We conduct object detection experiments on the well-known COCO 2017 benchmark~\cite{lin2014microsoft}. COCO 2017 dataset consists of 118K images for training and 5K images for validation, with annotations provided for 80 object categories. To evaluate the effectiveness of TCFormer, we utilize it as the backbone of both RetinaNet~\cite{lin2017focal} and Mask-RCNN~\cite{he2017mask}. 
Additionally, we replace the FPN module in both RetinaNet and Mask-RCNN frameworks with our MTA module to fully utilize the advantages of our dynamic tokens.
We follow most of the default settings of  MMDetection~\cite{mmdetection}, with the exception that we optimize the models using an AdamW optimizer with an initial learning rate of $1 \times 10^{-4}$. We adopt the common $1\times$ and $3\times$ setting for fair comparisons.

\textbf{Results.}
Qualitative results of TCFormer with Mask R-CNN are given in Fig.~\ref{fig:det_vis}.
%
We provide qualitative results of TCFormer with Mask R-CNN in Fig.~\ref{fig:det_vis}. As shown in Table~\ref{table:det}, TCFormer outperforms its counterparts on the RetinaNet framework. Specifically, the mAP of TCFormerV2-Small is $3.5$ points higher than Swin-T~\cite{liu2021swin} and $4.6$ points higher than PVT-Small~\cite{wang2021pyramid}. Compared with the vision transformers generating grid-base vision tokens, the advantage of TCFormer is particularly significant in the detection of small objects. For example, TCFormerV2-Small surpasses PVTv2-B2~\cite{wang2022pvt} by $1.4\%$ AP on the metric for small object detection (AP$_{\rm S}$), while the improvement on the overall metric is $0.4\%$ AP. This demonstrates that the dynamic vision tokens used in TCFormer are effective in capturing image details of small size, in line with our goals.

In Table~\ref{table:det}, we report the results of object detection and instance segmentation on the Mask R-CNN framework. The results indicate that TCFormer achieves superior performance to its counterparts. For instance, under $1 \times$ setting, TCFormerV2-Small outperforms Swin-T by $3.9\%$ AP on object detection and $2.8\%$ AP on instance segmentation. Similarly, when compared to PVTv2-B2, TCFormerV2-Small achieves an improvement of $0.8\%$ AP on object detection and $0.7\%$ AP on instance segmentation. 
The consistent improvements observed in both the RetinaNet and Mask R-CNN frameworks serve as evidence of the general strength and effectiveness of our TCFormer.

\section{Analysis}

\subsection{Ablation Study}

\textbf{Model Components.} We conduct ablative analysis on the semantic segmentation task using the ADE20K benchmark. We incorporate the TCFormerV2-Small into the Semantic FPN framework and present the results in Table~\ref{tab:ablation}. The experimental setup remains consistent with that outlined in Section~\ref{sec:seg}. 

The notable improvement of both the global CTM module ($0.9\%$ increase in mIoU) and the local CTM module ($0.5\%$ increase in mIoU) confirms the benefits of our dynamic vision tokens in image feature learning. However, the high complexity involved by the global CTM module ($42.1$ GFLOPs) makes it impractical. In contrast, the local CTM module is considerably more efficient and only involves only $5.6$ GFLOPs, which is $86.7\%$ less than the global CTM module.

As previously discussed in Section~\ref{sec:tcformer:mta}, traditional CNN-based feature aggregation modules fail to fully exploit the details captured by our dynamic vision tokens. As an evidence of our opinion, replacing the FPN module with our MTA module results in further enhancements, which demonstrates the effectiveness of our MTA module design.
Compared to the original SR-MTA module, the CR-MTA module is both more effective and more efficient. CR-MTA module brings less parameters and complexity, but presents a significant performance gain over the SR-MTA module: $0.5\%$ mIoU when using global CTM modules and $0.7\%$ mIoU when using local CTM modules. The experimental results prove that the CR-MTA module can fully leverage the benefits of our dynamic tokens, as illustrated in Section~\ref{sec:tcformer:cr_mta}.

\textbf{Clustering Methods.}
We assess the impact of different clustering methods by integrating them into the TCFormerV2-Small model and evaluating their performance on the ImageNet validation set. The results are detailed in Table~\ref{tab:ablation_cluster}. 
For the K-means algorithm, we follow ~\cite{liang2022expediting} to initialize the cluster centers using adaptive average pooling on the feature map and then update the clustering results through $10$ iterations. The Bipartite Soft Matching (BSF) method~\cite{bolya2022token} divides tokens into two sets and merges token pairs with the highest similarity between these sets. Since a single BSF step cannot reduce the token number to a quarter of the original, we employ $5$ BSF steps at the end of each stage to achieve the same token reduction as other methods.
It is important to note that we do not fine-tune the models further after changing the clustering method. 

As demonstrated in Table~\ref{tab:ablation_cluster}, our model exhibits robustness to different clustering methods. Without additional fine-tuning, the model trained with the local DPC-kNN algorithm adapts well to other clustering methods, showing negligible performance drops. Ultimately, we select the local DPC-kNN algorithm for its efficiency.


\begin{table}[tb]
\centering
\setlength{\tabcolsep}{1.5mm}
\caption{\textbf{Components ablation study on semantic segmentation on the ADE20K validation set.}
    ``GFLOPs'' is calculated under the input scale of $512\times 512$.
}
\label{tab:ablation}
\begin{tabular}{c|c|c|c|c}
\Thline
\renewcommand{\arraystretch}{0.1}
\makecell{Token Merge \\ Module} & \makecell{Feature Aggregation \\ Module} & \#Param & GFLOPs & mIoU$\%$ \\
\hline
CNN & FPN & $29.1M$ & $45.8$ & $45.2$ \\
Global CTM & FPN & $29.4M$ & $87.9$ & $46.1$ \\
Global CTM & SR-MTA & $33.7M$ & $85.1$ & $46.5$ \\
Global CTM & CR-MTA & $28.4M$ & $80.9$ & $47.0$ \\
Local CTM & FPN & $29.4M$ & $51.4$ & $45.7$ \\
Local CTM & SR-MTA & $33.7M$ & $48.6$ & $47.1$ \\
\rowcolor{Gray}
Local CTM & CR-MTA & $28.4M$ & $44.4$ & $47.8$ \\
\Thline
\end{tabular}
\end{table}

\begin{table}[tb]
\centering
\setlength{\tabcolsep}{1.5mm}
\caption{
\textbf{Comparisons between clustering methods on image classification on ImageNet validation set.}
Various clustering methods are applied to the TCFormerV2-Small model, which is trained using the local DPC-kNN algorithm without additional fine-tuning. ``GFLOPs'' is calculated under the input scale of $224\times 224$. Throughput is measured using the GitHub repository of~\cite{rw2019timm} and a V100 GPU.
}
\label{tab:ablation_cluster}
\begin{tabular}{c|c|c|c}
\Thline
\renewcommand{\arraystretch}{0.1}
 \makecell{Clustering Method } & GFLOPs & \makecell{Throughput\\(images/s)} & \makecell{Top-1 Acc} \\
\hline
K-means &$11.7$ & $114.7$ & $82.39$ \\
Bipartite Soft Matching~\cite{bolya2022token} & $6.0$ & $207.5$ & $82.31$ \\
DPC-kNN & $5.9$ & $120.8$ & $82.44$ \\
\rowcolor{Gray}
Local DPC-kNN & $4.5$ & $237.5$ & $82.40$ \\
\Thline
\end{tabular}
\end{table}

\begin{figure}[tb]
	\centering
	\includegraphics[width=0.48\textwidth]{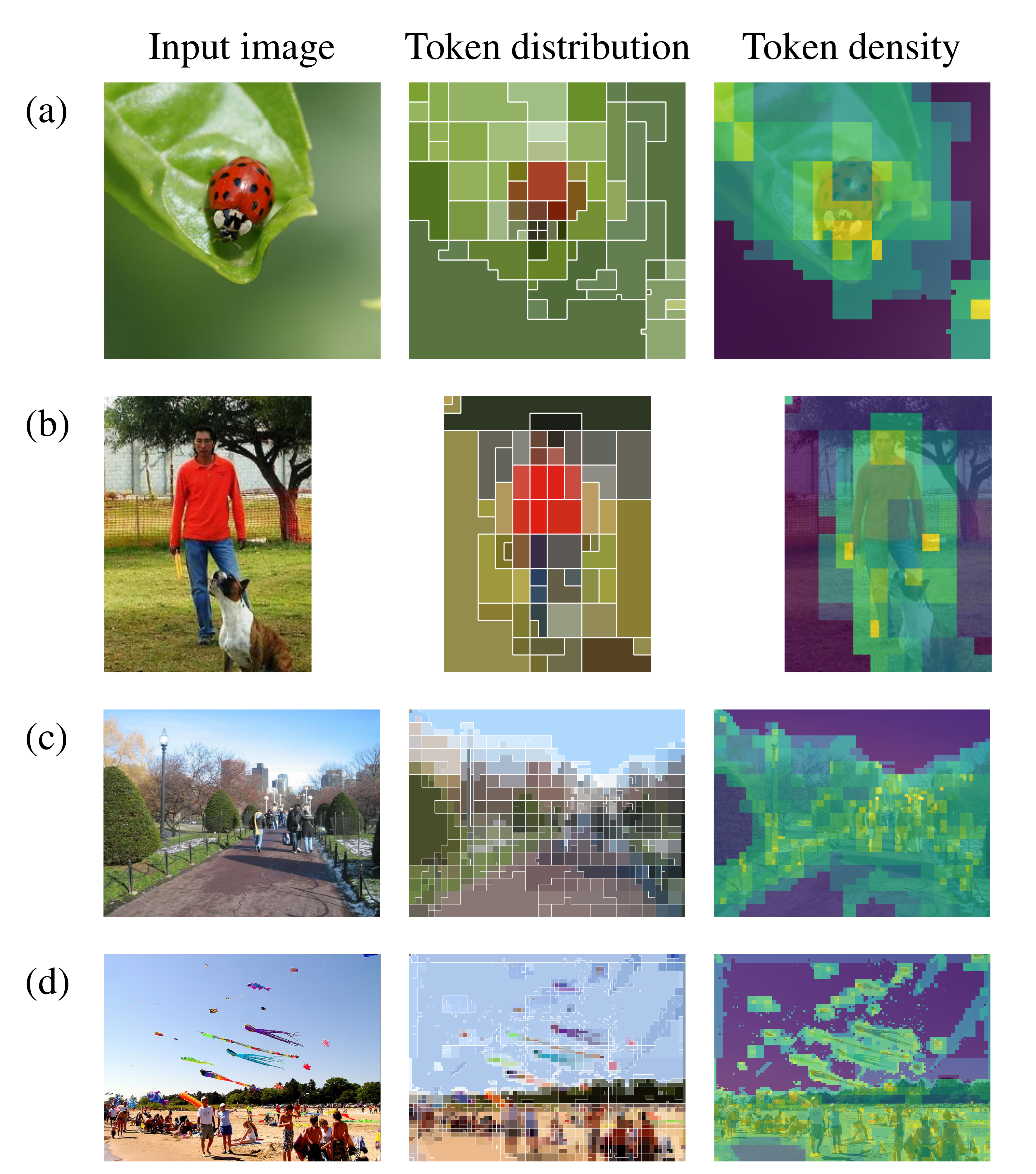}
        \caption{
        Examples of the dynamic vision tokens generated by TCFormer on (a) image classification, (b) human pose estimation, (c) semantic segmentation and (d) object detection.
        }
	\label{fig:token_vis}
\end{figure}

\begin{figure}[tb]
	\centering
	\includegraphics[width=0.41\textwidth]{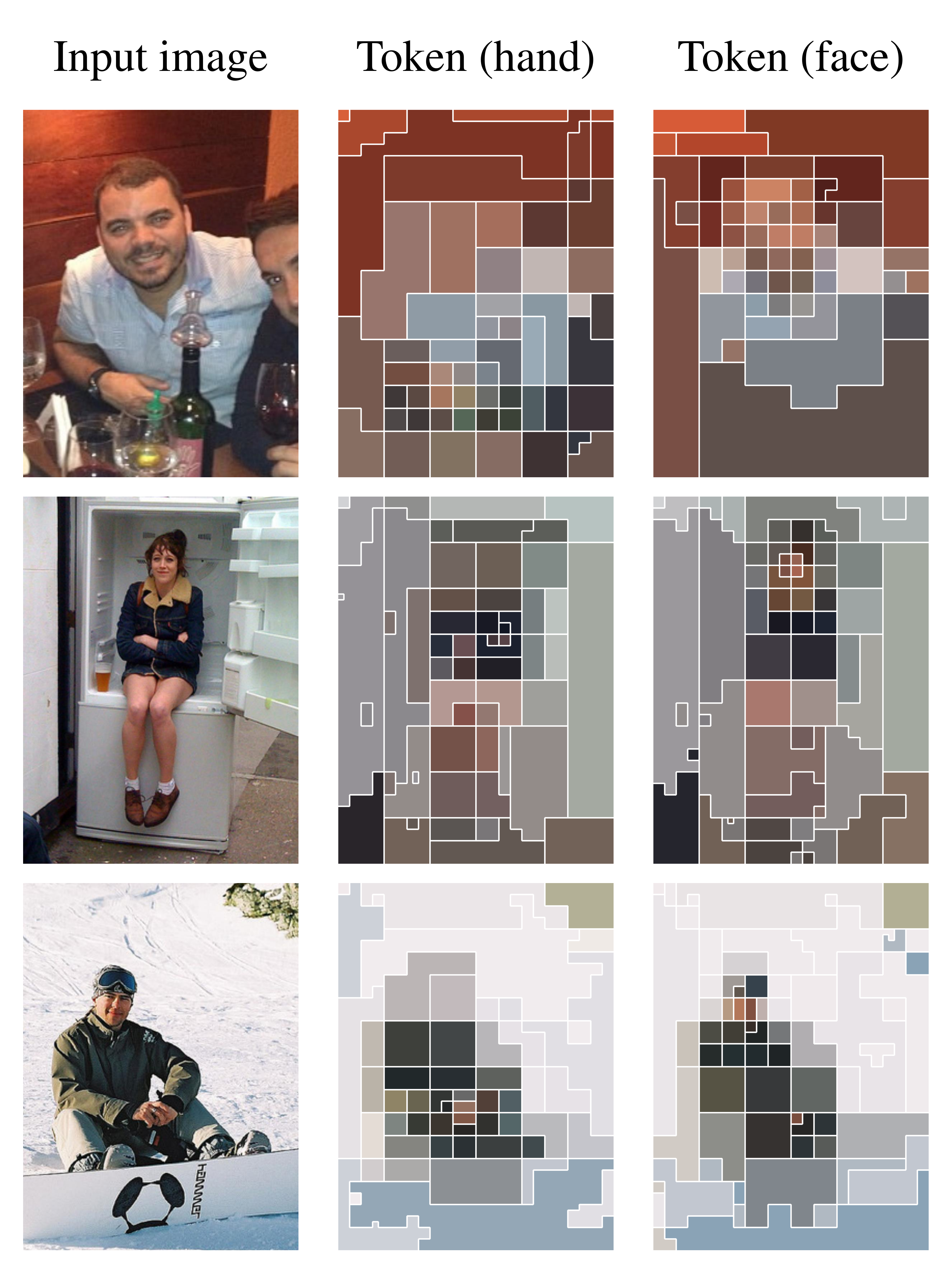}
        \caption{
        Examples of the dynamic tokens for different tasks. We show the input image and the dynamic tokens for hand keypoint estimation (Token (hand)) and face keypoint estimation (Token (face)). TCFormer adjusts the dynamic token according to the task.
        }
	\label{fig:token_compare}
\end{figure}

\subsection{Token Distribution}
Fig.~\ref{fig:token_vis} shows some examples of the dynamic vision tokens generated by TCFormer across various tasks, including image classification, human pose estimation, semantic segmentation, and object detection. The advantages of our dynamic vision tokens are demonstrated in the examples.

%
Firstly, as shown in Fig.~\ref{fig:token_vis}, the dynamic tokens generated by TCFormer are well-aligned with the objects in the input images. Such alignment provides a clearer semantic meaning compared to traditional fixed grid-based tokens, which facilitates the learning of both token features and object relationships. 
Secondly, TCFormer utilizes fine tokens to represent intricate details in small sizes, such as the human hands in Fig.~\ref{fig:token_vis}(b) and the kite in Fig.~\ref{fig:token_vis}(d). Representing such details with fine tokens allows TCFormer to capture detailed information more effectively.

Thirdly, TCFormer adjusts the token distribution in accordance with the task at hand and allocates more tokens to the crucial regions, thus enabling the model to focus on these regions and learn a better representation of the image. 
For image classification and human pose estimation tasks, there is typically a distinct subject in the input image. As shown in Fig.~\ref{fig:token_vis}(a) and Fig.~\ref{fig:token_vis}(b),  TCFormer distinguishes the background regions from the regions belonging to the subject and allocates most of the tokens to the latter. Conversely, the background regions are represented by only a few vision tokens. 
In the case of semantic segmentation and object detection tasks, where there is usually no clear subject, TCFormer adjusts the token distribution based on the amount of information present in the image regions. As demonstrated in in Fig.~\ref{fig:token_vis}(c) and Fig.~\ref{fig:token_vis}(d), TCFormer represents simple regions with fewer tokens and allocates more tokens to the complex regions. 
It is worth noting that even complex backgrounds do not consume more tokens in image classification and human pose estimation tasks, such as the example in Fig.~\ref{fig:token_vis}(b), as the information of the background is not valuable to these tasks. 
To further support our conclusion, we train two models to estimate human hand and face keypoints, respectively. The dynamic tokens generated by these two models are displayed in Fig.~\ref{fig:token_compare}. The task-specific dynamic tokens demonstrate that TCFormer can automatically adjust the token distribution for different tasks and focus on the crucial regions.

\section{Conclusion}
\label{sec:conclusion}
In this paper, we present the Token Clustering Transformer (TCFormer), a novel transformer-based architecture suitable for a wide spectrum of vision tasks. 
TCFormer generates dynamic tokens that enhance the model's ability to focus on crucial regions and preserve intricate details while disregarding unnecessary background information. 
Our extensive experiments across multiple vision tasks, including image classification, human pose estimation, semantic segmentation, and object detection, show that TCFormer outperforms state-of-the-art transformer-based backbones while maintaining comparable parameter numbers. We believe that TCFormer has the potential to be utilized in a multitude of applications. We hope our research can inspire further exploration in the realm of dynamic vision tokens, leading to the development of more advanced architectures.

\ifCLASSOPTIONcompsoc
  \section*{Acknowledgments}
  This project is funded in part by the Centre for Perceptual and Interactive Intelligence (CPIl) Ltd under the Innovation and Technology Commission (ITC)’s InnoHK, by Research Impact Fund Project R5001-18 of Hong Kong RGC. Xiaogang Wang is a PI of CPII under the InnoHK.
  
\else
  \section*{Acknowledgment}
\fi



\ifCLASSOPTIONcaptionsoff
  \newpage
\fi


\bibliographystyle{IEEEtran}
\bibliography{bib/ref}

\begin{thebibliography}{10}
\providecommand{\url}[1]{#1}
\csname url@samestyle\endcsname
\providecommand{\newblock}{\relax}
\providecommand{\bibinfo}[2]{#2}
\providecommand{\BIBentrySTDinterwordspacing}{\spaceskip=0pt\relax}
\providecommand{\BIBentryALTinterwordstretchfactor}{4}
\providecommand{\BIBentryALTinterwordspacing}{\spaceskip=\fontdimen2\font plus
\BIBentryALTinterwordstretchfactor\fontdimen3\font minus \fontdimen4\font\relax}
\providecommand{\BIBforeignlanguage}[2]{{%
\expandafter\ifx\csname l@#1\endcsname\relax
\typeout{** WARNING: IEEEtran.bst: No hyphenation pattern has been}%
\typeout{** loaded for the language `#1'. Using the pattern for}%
\typeout{** the default language instead.}%
\else
\language=\csname l@#1\endcsname
\fi
#2}}
\providecommand{\BIBdecl}{\relax}
\BIBdecl

\bibitem{dosovitskiy2020image}
A.~Dosovitskiy, L.~Beyer, A.~Kolesnikov, D.~Weissenborn, X.~Zhai, T.~Unterthiner, M.~Dehghani, M.~Minderer, G.~Heigold, S.~Gelly \emph{et~al.}, ``An image is worth 16x16 words: Transformers for image recognition at scale,'' in \emph{Int. Conf. Learn. Represent.}, 2020.

\bibitem{touvron2021training}
H.~Touvron, M.~Cord, M.~Douze, F.~Massa, A.~Sablayrolles, and H.~J{\'e}gou, ``Training data-efficient image transformers \& distillation through attention,'' in \emph{ICML}.\hskip 1em plus 0.5em minus 0.4em\relax PMLR, 2021.

\bibitem{touvron2021going}
H.~Touvron, M.~Cord, A.~Sablayrolles, G.~Synnaeve, and H.~J{\'e}gou, ``Going deeper with image transformers,'' in \emph{Int. Conf. Comput. Vis.}, 2021.

\bibitem{wu2021cvt}
H.~Wu, B.~Xiao, N.~Codella, M.~Liu, X.~Dai, L.~Yuan, and L.~Zhang, ``Cvt: Introducing convolutions to vision transformers,'' in \emph{Int. Conf. Comput. Vis.}, 2021.

\bibitem{carion2020end}
N.~Carion, F.~Massa, G.~Synnaeve, N.~Usunier, A.~Kirillov, and S.~Zagoruyko, ``End-to-end object detection with transformers,'' in \emph{Eur. Conf. Comput. Vis.}, 2020.

\bibitem{zhu2020deformable}
X.~Zhu, W.~Su, L.~Lu, B.~Li, X.~Wang, and J.~Dai, ``Deformable detr: Deformable transformers for end-to-end object detection,'' in \emph{Int. Conf. Learn. Represent.}, 2020.

\bibitem{zhang2022dino}
H.~Zhang, F.~Li, S.~Liu, L.~Zhang, H.~Su, J.~Zhu, L.~M. Ni, and H.-Y. Shum, ``Dino: Detr with improved denoising anchor boxes for end-to-end object detection,'' \emph{arXiv preprint arXiv:2203.03605}, 2022.

\bibitem{liu2021swin}
Z.~Liu, Y.~Lin, Y.~Cao, H.~Hu, Y.~Wei, Z.~Zhang, S.~Lin, and B.~Guo, ``Swin transformer: Hierarchical vision transformer using shifted windows,'' in \emph{Int. Conf. Comput. Vis.}, 2021.

\bibitem{xie2021segformer}
E.~Xie, W.~Wang, Z.~Yu, A.~Anandkumar, J.~M. Alvarez, and P.~Luo, ``Segformer: Simple and efficient design for semantic segmentation with transformers,'' \emph{Adv. Neural Inform. Process. Syst.}, vol.~34, 2021.

\bibitem{cheng2022masked}
B.~Cheng, I.~Misra, A.~G. Schwing, A.~Kirillov, and R.~Girdhar, ``Masked-attention mask transformer for universal image segmentation,'' in \emph{IEEE Conf. Comput. Vis. Pattern Recog.}, 2022.

\bibitem{strudel2021segmenter}
R.~Strudel, R.~Garcia, I.~Laptev, and C.~Schmid, ``Segmenter: Transformer for semantic segmentation,'' in \emph{Int. Conf. Comput. Vis.}, 2021.

\bibitem{chen2022vision}
Z.~Chen, Y.~Duan, W.~Wang, J.~He, T.~Lu, J.~Dai, and Y.~Qiao, ``Vision transformer adapter for dense predictions,'' \emph{arXiv preprint arXiv:2205.08534}, 2022.

\bibitem{yuan2021hrformer}
Y.~Yuan, R.~Fu, L.~Huang, W.~Lin, C.~Zhang, X.~Chen, and J.~Wang, ``Hrformer: High-resolution vision transformer for dense predict,'' \emph{Adv. Neural Inform. Process. Syst.}, vol.~34, 2021.

\bibitem{xu2022vitpose}
Y.~Xu, J.~Zhang, Q.~Zhang, and D.~Tao, ``Vitpose: Simple vision transformer baselines for human pose estimation,'' \emph{arXiv preprint arXiv:2204.12484}, 2022.

\bibitem{zeng2022not}
W.~Zeng, S.~Jin, W.~Liu, C.~Qian, P.~Luo, W.~Ouyang, and X.~Wang, ``Not all tokens are equal: Human-centric visual analysis via token clustering transformer,'' in \emph{Proceedings of the IEEE/CVF Conference on Computer Vision and Pattern Recognition}, 2022, pp. 11\,101--11\,111.

\bibitem{wang2022pvt}
W.~Wang, E.~Xie, X.~Li, D.-P. Fan, K.~Song, D.~Liang, T.~Lu, P.~Luo, and L.~Shao, ``Pvt v2: Improved baselines with pyramid vision transformer,'' \emph{Computational Visual Media}, vol.~8, no.~3, pp. 415--424, 2022.

\bibitem{meng2022adavit}
L.~Meng, H.~Li, B.-C. Chen, S.~Lan, Z.~Wu, Y.-G. Jiang, and S.-N. Lim, ``Adavit: Adaptive vision transformers for efficient image recognition,'' in \emph{IEEE Conf. Comput. Vis. Pattern Recog.}, 2022.

\bibitem{sun2019deep}
K.~Sun, B.~Xiao, D.~Liu, and J.~Wang, ``Deep high-resolution representation learning for human pose estimation,'' in \emph{IEEE Conf. Comput. Vis. Pattern Recog.}, 2019.

\bibitem{lin2017feature}
T.-Y. Lin, P.~Doll{\'a}r, R.~Girshick, K.~He, B.~Hariharan, and S.~Belongie, ``Feature pyramid networks for object detection,'' in \emph{IEEE Conf. Comput. Vis. Pattern Recog.}, 2017.

\bibitem{wang2021pyramid}
W.~Wang, E.~Xie, X.~Li, D.-P. Fan, K.~Song, D.~Liang, T.~Lu, P.~Luo, and L.~Shao, ``Pyramid vision transformer: A versatile backbone for dense prediction without convolutions,'' in \emph{Int. Conf. Comput. Vis.}, 2021.

\bibitem{he2022masked}
K.~He, X.~Chen, S.~Xie, Y.~Li, P.~Doll{\'a}r, and R.~Girshick, ``Masked autoencoders are scalable vision learners,'' in \emph{IEEE Conf. Comput. Vis. Pattern Recog.}, 2022.

\bibitem{rao2021dynamicvit}
Y.~Rao, W.~Zhao, B.~Liu, J.~Lu, J.~Zhou, and C.-J. Hsieh, ``Dynamicvit: Efficient vision transformers with dynamic token sparsification,'' \emph{Adv. Neural Inform. Process. Syst.}, 2021.

\bibitem{wang2021pnp}
T.~Wang, L.~Yuan, Y.~Chen, J.~Feng, and S.~Yan, ``Pnp-detr: Towards efficient visual analysis with transformers,'' in \emph{Int. Conf. Comput. Vis.}, 2021, pp. 4661--4670.

\bibitem{xu2022evo}
Y.~Xu, Z.~Zhang, M.~Zhang, K.~Sheng, K.~Li, W.~Dong, L.~Zhang, C.~Xu, and X.~Sun, ``Evo-vit: Slow-fast token evolution for dynamic vision transformer,'' in \emph{AAAI}, 2022.

\bibitem{wang2021not}
Y.~Wang, R.~Huang, S.~Song, Z.~Huang, and G.~Huang, ``Not all images are worth 16x16 words: Dynamic transformers for efficient image recognition,'' \emph{Adv. Neural Inform. Process. Syst.}, 2021.

\bibitem{yue2021vision}
X.~Yue, S.~Sun, Z.~Kuang, M.~Wei, P.~H. Torr, W.~Zhang, and D.~Lin, ``Vision transformer with progressive sampling,'' in \emph{Int. Conf. Comput. Vis.}, 2021.

\bibitem{bolya2022token}
D.~Bolya, C.-Y. Fu, X.~Dai, P.~Zhang, C.~Feichtenhofer, and J.~Hoffman, ``Token merging: Your vit but faster,'' \emph{arXiv preprint arXiv:2210.09461}, 2022.

\bibitem{liang2022expediting}
W.~Liang, Y.~Yuan, H.~Ding, X.~Luo, W.~Lin, D.~Jia, Z.~Zhang, C.~Zhang, and H.~Hu, ``Expediting large-scale vision transformer for dense prediction without fine-tuning,'' \emph{Advances in Neural Information Processing Systems}, vol.~35, pp. 35\,462--35\,477, 2022.

\bibitem{li2021localvit}
Y.~Li, K.~Zhang, J.~Cao, R.~Timofte, and L.~Van~Gool, ``Localvit: Bringing locality to vision transformers,'' \emph{arXiv preprint arXiv:2104.05707}, 2021.

\bibitem{du2016study}
M.~Du, S.~Ding, and H.~Jia, ``Study on density peaks clustering based on k-nearest neighbors and principal component analysis,'' \emph{Knowledge-Based Systems}, vol.~99, pp. 135--145, 2016.

\bibitem{rw2019timm}
R.~Wightman, ``Pytorch image models,'' \url{https://github.com/rwightman/pytorch-image-models}, 2019.

\bibitem{he2016deep}
K.~He, X.~Zhang, S.~Ren, and J.~Sun, ``Deep residual learning for image recognition,'' in \emph{IEEE Conf. Comput. Vis. Pattern Recog.}, 2016.

\bibitem{xie2017aggregated}
S.~Xie, R.~Girshick, P.~Doll{\'a}r, Z.~Tu, and K.~He, ``Aggregated residual transformations for deep neural networks,'' in \emph{Proceedings of the IEEE conference on computer vision and pattern recognition}, 2017, pp. 1492--1500.

\bibitem{radosavovic2020designing}
I.~Radosavovic, R.~P. Kosaraju, R.~Girshick, K.~He, and P.~Doll{\'a}r, ``Designing network design spaces,'' in \emph{Proceedings of the IEEE/CVF conference on computer vision and pattern recognition}, 2020, pp. 10\,428--10\,436.

\bibitem{yuan2021tokens}
L.~Yuan, Y.~Chen, T.~Wang, W.~Yu, Y.~Shi, Z.-H. Jiang, F.~E. Tay, J.~Feng, and S.~Yan, ``Tokens-to-token vit: Training vision transformers from scratch on imagenet,'' in \emph{Proceedings of the IEEE/CVF international conference on computer vision}, 2021, pp. 558--567.

\bibitem{han2021transformer}
K.~Han, A.~Xiao, E.~Wu, J.~Guo, C.~Xu, and Y.~Wang, ``Transformer in transformer,'' \emph{Adv. Neural Inform. Process. Syst.}, 2021.

\bibitem{xu2021co}
W.~Xu, Y.~Xu, T.~Chang, and Z.~Tu, ``Co-scale conv-attentional image transformers,'' in \emph{Proceedings of the IEEE/CVF International Conference on Computer Vision}, 2021, pp. 9981--9990.

\bibitem{chu2021twins}
X.~Chu, Z.~Tian, Y.~Wang, B.~Zhang, H.~Ren, X.~Wei, H.~Xia, and C.~Shen, ``Twins: Revisiting the design of spatial attention in vision transformers,'' \emph{Advances in Neural Information Processing Systems}, vol.~34, pp. 9355--9366, 2021.

\bibitem{si2022inception}
C.~Si, W.~Yu, P.~Zhou, Y.~Zhou, X.~Wang, and S.~Yan, ``Inception transformer,'' \emph{Advances in Neural Information Processing Systems}, vol.~35, pp. 23\,495--23\,509, 2022.

\bibitem{russakovsky2015imagenet}
O.~Russakovsky, J.~Deng, H.~Su, J.~Krause, S.~Satheesh, S.~Ma, Z.~Huang, A.~Karpathy, A.~Khosla, M.~Bernstein \emph{et~al.}, ``Imagenet large scale visual recognition challenge,'' \emph{International journal of computer vision}, vol. 115, no.~3, pp. 211--252, 2015.

\bibitem{szegedy2015going}
C.~Szegedy, W.~Liu, Y.~Jia, P.~Sermanet, S.~Reed, D.~Anguelov, D.~Erhan, V.~Vanhoucke, and A.~Rabinovich, ``Going deeper with convolutions,'' in \emph{IEEE Conf. Comput. Vis. Pattern Recog.}, 2015.

\bibitem{szegedy2016rethinking}
C.~Szegedy, V.~Vanhoucke, S.~Ioffe, J.~Shlens, and Z.~Wojna, ``Rethinking the inception architecture for computer vision,'' in \emph{IEEE Conf. Comput. Vis. Pattern Recog.}, 2016.

\bibitem{zhang2017mixup}
H.~Zhang, M.~Cisse, Y.~N. Dauphin, and D.~Lopez-Paz, ``mixup: Beyond empirical risk minimization,'' \emph{Int. Conf. Learn. Represent.}, 2018.

\bibitem{yun2019cutmix}
S.~Yun, D.~Han, S.~J. Oh, S.~Chun, J.~Choe, and Y.~Yoo, ``Cutmix: Regularization strategy to train strong classifiers with localizable features,'' in \emph{Int. Conf. Comput. Vis.}, 2019, pp. 6023--6032.

\bibitem{zhong2020random}
Z.~Zhong, L.~Zheng, G.~Kang, S.~Li, and Y.~Yang, ``Random erasing data augmentation,'' in \emph{AAAI}, 2020.

\bibitem{loshchilov2018decoupled}
I.~Loshchilov and F.~Hutter, ``Decoupled weight decay regularization,'' in \emph{International Conference on Learning Representations}, 2018.

\bibitem{loshchilov2016sgdr}
------, ``Sgdr: Stochastic gradient descent with warm restarts,'' \emph{Int. Conf. Learn. Represent.}, 2017.

\bibitem{mmpose2020}
M.~Contributors, ``Openmmlab pose estimation toolbox and benchmark,'' \url{https://github.com/open-mmlab/mmpose}, 2020.

\bibitem{hidalgo2019single}
G.~Hidalgo, Y.~Raaj, H.~Idrees, D.~Xiang, H.~Joo, T.~Simon, and Y.~Sheikh, ``Single-network whole-body pose estimation,'' in \emph{IEEE Conf. Comput. Vis. Pattern Recog.}, 2019.

\bibitem{cao2018openpose}
Z.~Cao, G.~Hidalgo, T.~Simon, S.-E. Wei, and Y.~Sheikh, ``Openpose: realtime multi-person 2d pose estimation using part affinity fields,'' \emph{IEEE Trans. Pattern Anal. Mach. Intell.}, 2018.

\bibitem{cao2017realtime}
Z.~Cao, T.~Simon, S.-E. Wei, and Y.~Sheikh, ``Realtime multi-person 2d pose estimation using part affinity fields,'' in \emph{IEEE Conf. Comput. Vis. Pattern Recog.}, 2017.

\bibitem{newell2017associative}
A.~Newell, Z.~Huang, and J.~Deng, ``Associative embedding: End-to-end learning for joint detection and grouping,'' in \emph{Adv. Neural Inform. Process. Syst.}, 2017.

\bibitem{cheng2020higherhrnet}
B.~Cheng, B.~Xiao, J.~Wang, H.~Shi, T.~S. Huang, and L.~Zhang, ``Higherhrnet: Scale-aware representation learning for bottom-up human pose estimation,'' in \emph{IEEE Conf. Comput. Vis. Pattern Recog.}, 2020, pp. 5386--5395.

\bibitem{xiao2018simple}
B.~Xiao, H.~Wu, and Y.~Wei, ``Simple baselines for human pose estimation and tracking,'' in \emph{Eur. Conf. Comput. Vis.}, 2018.

\bibitem{jin2020whole}
S.~Jin, L.~Xu, J.~Xu, C.~Wang, W.~Liu, C.~Qian, W.~Ouyang, and P.~Luo, ``Whole-body human pose estimation in the wild,'' in \emph{Eur. Conf. Comput. Vis.}, 2020, pp. 196--214.

\bibitem{xu2022zoomnas}
L.~Xu, S.~Jin, W.~Liu, C.~Qian, W.~Ouyang, P.~Luo, and X.~Wang, ``Zoomnas: searching for whole-body human pose estimation in the wild,'' \emph{IEEE Transactions on Pattern Analysis and Machine Intelligence}, 2022.

\bibitem{lin2014microsoft}
T.-Y. Lin, M.~Maire, S.~Belongie, J.~Hays, P.~Perona, D.~Ramanan, P.~Doll{\'a}r, and C.~L. Zitnick, ``Microsoft coco: Common objects in context,'' in \emph{Eur. Conf. Comput. Vis.}, 2014.

\bibitem{kingma2014adam}
D.~P. Kingma and J.~Ba, ``Adam: A method for stochastic optimization,'' \emph{Int. Conf. Learn. Represent.}, 2015.

\bibitem{newell2016stacked}
A.~Newell, K.~Yang, and J.~Deng, ``Stacked hourglass networks for human pose estimation,'' in \emph{Computer Vision--ECCV 2016: 14th European Conference, Amsterdam, The Netherlands, October 11-14, 2016, Proceedings, Part VIII 14}.\hskip 1em plus 0.5em minus 0.4em\relax Springer, 2016, pp. 483--499.

\bibitem{chen2018cascaded}
Y.~Chen, Z.~Wang, Y.~Peng, Z.~Zhang, G.~Yu, and J.~Sun, ``Cascaded pyramid network for multi-person pose estimation,'' in \emph{Proceedings of the IEEE conference on computer vision and pattern recognition}, 2018, pp. 7103--7112.

\bibitem{toshev2014deeppose}
A.~Toshev and C.~Szegedy, ``Deeppose: Human pose estimation via deep neural networks,'' in \emph{Proceedings of the IEEE conference on computer vision and pattern recognition}, 2014, pp. 1653--1660.

\bibitem{yang2021transpose}
S.~Yang, Z.~Quan, M.~Nie, and W.~Yang, ``Transpose: Keypoint localization via transformer,'' in \emph{IEEE/CVF International Conference on Computer Vision (ICCV)}, 2021.

\bibitem{li2021human}
J.~Li, S.~Bian, A.~Zeng, C.~Wang, B.~Pang, W.~Liu, and C.~Lu, ``Human pose regression with residual log-likelihood estimation,'' in \emph{Int. Conf. Comput. Vis.}, 2021, pp. 11\,025--11\,034.

\bibitem{jin2020differentiable}
S.~Jin, W.~Liu, E.~Xie, W.~Wang, C.~Qian, W.~Ouyang, and P.~Luo, ``Differentiable hierarchical graph grouping for multi-person pose estimation,'' in \emph{Eur. Conf. Comput. Vis.}\hskip 1em plus 0.5em minus 0.4em\relax Springer, 2020, pp. 718--734.

\bibitem{he2017mask}
K.~He, G.~Gkioxari, P.~Doll{\'a}r, and R.~Girshick, ``Mask r-cnn,'' in \emph{Proceedings of the IEEE international conference on computer vision}, 2017, pp. 2961--2969.

\bibitem{papandreou2017towards}
G.~Papandreou, T.~Zhu, N.~Kanazawa, A.~Toshev, J.~Tompson, C.~Bregler, and K.~Murphy, ``Towards accurate multi-person pose estimation in the wild,'' in \emph{Proceedings of the IEEE conference on computer vision and pattern recognition}, 2017, pp. 4903--4911.

\bibitem{sun2018integral}
X.~Sun, B.~Xiao, F.~Wei, S.~Liang, and Y.~Wei, ``Integral human pose regression,'' in \emph{Proceedings of the European conference on computer vision (ECCV)}, 2018, pp. 529--545.

\bibitem{li2021tokenpose}
Y.~Li, S.~Zhang, Z.~Wang, S.~Yang, W.~Yang, S.-T. Xia, and E.~Zhou, ``Tokenpose: Learning keypoint tokens for human pose estimation,'' in \emph{Proceedings of the IEEE/CVF International conference on computer vision}, 2021, pp. 11\,313--11\,322.

\bibitem{zhou2019objects}
X.~Zhou, D.~Wang, and P.~Kr{\"a}henb{\"u}hl, ``Objects as points,'' \emph{arXiv preprint arXiv:1904.07850}, 2019.

\bibitem{wei2020point}
F.~Wei, X.~Sun, H.~Li, J.~Wang, and S.~Lin, ``Point-set anchors for object detection, instance segmentation and pose estimation,'' in \emph{European Conference on Computer Vision}.\hskip 1em plus 0.5em minus 0.4em\relax Springer, 2020, pp. 527--544.

\bibitem{li2021pose}
K.~Li, S.~Wang, X.~Zhang, Y.~Xu, W.~Xu, and Z.~Tu, ``Pose recognition with cascade transformers,'' in \emph{Proceedings of the IEEE/CVF Conference on Computer Vision and Pattern Recognition}, 2021, pp. 1944--1953.

\bibitem{zhou2017scene}
B.~Zhou, H.~Zhao, X.~Puig, S.~Fidler, A.~Barriuso, and A.~Torralba, ``Scene parsing through ade20k dataset,'' in \emph{Proceedings of the IEEE conference on computer vision and pattern recognition}, 2017, pp. 633--641.

\bibitem{kirillov2019panoptic}
A.~Kirillov, R.~Girshick, K.~He, and P.~Doll{\'a}r, ``Panoptic feature pyramid networks,'' in \emph{Proceedings of the IEEE/CVF conference on computer vision and pattern recognition}, 2019, pp. 6399--6408.

\bibitem{chen2021cyclemlp}
S.~Chen, E.~Xie, C.~Ge, D.~Liang, and P.~Luo, ``Cyclemlp: A mlp-like architecture for dense prediction,'' \emph{arXiv preprint arXiv:2107.10224}, 2021.

\bibitem{han2022vision}
K.~Han, Y.~Wang, J.~Guo, Y.~Tang, and E.~Wu, ``Vision gnn: An image is worth graph of nodes,'' \emph{arXiv preprint arXiv:2206.00272}, 2022.

\bibitem{yu2022metaformer}
W.~Yu, M.~Luo, P.~Zhou, C.~Si, Y.~Zhou, X.~Wang, J.~Feng, and S.~Yan, ``Metaformer is actually what you need for vision,'' in \emph{Proceedings of the IEEE/CVF conference on computer vision and pattern recognition}, 2022, pp. 10\,819--10\,829.

\bibitem{lin2017focal}
T.-Y. Lin, P.~Goyal, R.~Girshick, K.~He, and P.~Doll{\'a}r, ``Focal loss for dense object detection,'' in \emph{Proceedings of the IEEE international conference on computer vision}, 2017, pp. 2980--2988.

\bibitem{mmdetection}
K.~Chen, J.~Wang, J.~Pang, Y.~Cao, Y.~Xiong, X.~Li, S.~Sun, W.~Feng, Z.~Liu, J.~Xu, Z.~Zhang, D.~Cheng, C.~Zhu, T.~Cheng, Q.~Zhao, B.~Li, X.~Lu, R.~Zhu, Y.~Wu, J.~Dai, J.~Wang, J.~Shi, W.~Ouyang, C.~C. Loy, and D.~Lin, ``{MMDetection}: Open mmlab detection toolbox and benchmark,'' \emph{arXiv preprint arXiv:1906.07155}, 2019.

\end{thebibliography}

\clearpage

\begin{IEEEbiography}[{\includegraphics[width=1in,height=1.25in,clip,keepaspectratio]{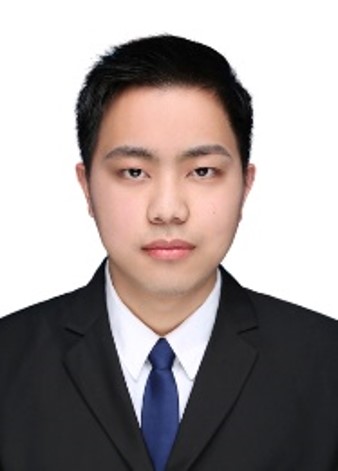}}]{Wang Zeng}
received the B.Eng. degree from the College of Optical Science and Engineering, Zhejiang University, Hangzhou, China, in 2018. He is currently a Ph.D. candidate with the Department of Electronic Engineering, The Chinese University of Hong Kong, Hong Kong SAR, China. His research interests include computer vision, deep learning, and human pose estimation.
\end{IEEEbiography}

\begin{IEEEbiography}[{\includegraphics[width=1in,height=1.25in,clip,keepaspectratio]{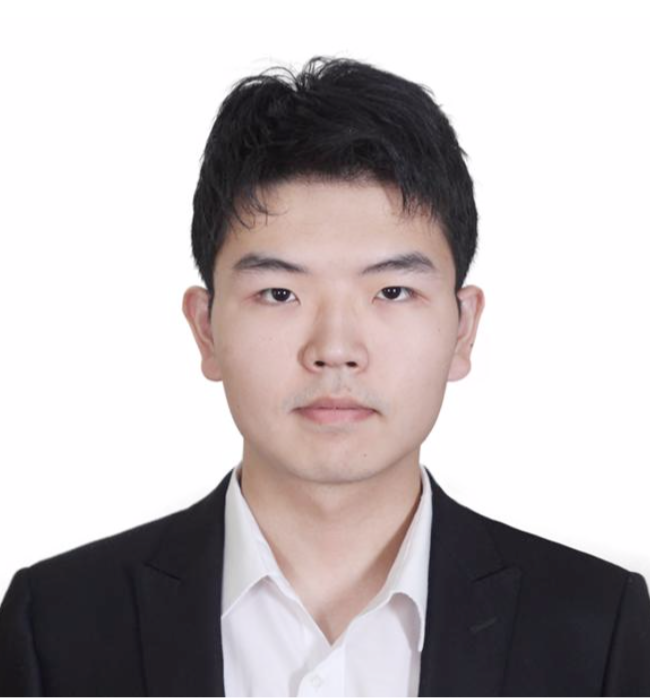}}]{Sheng Jin}
received the B.Eng. and M.Eng. degrees from the Department of Automation, Tsinghua University, Beijing, China, in 2017 and 2020. He is currently a Ph.D. student at the University of Hong Kong, Hong Kong SAR, China. His research interests include deep learning and human pose estimation.
\end{IEEEbiography}

\begin{IEEEbiography}[{\includegraphics[width=1in,height=1.25in,clip,keepaspectratio]{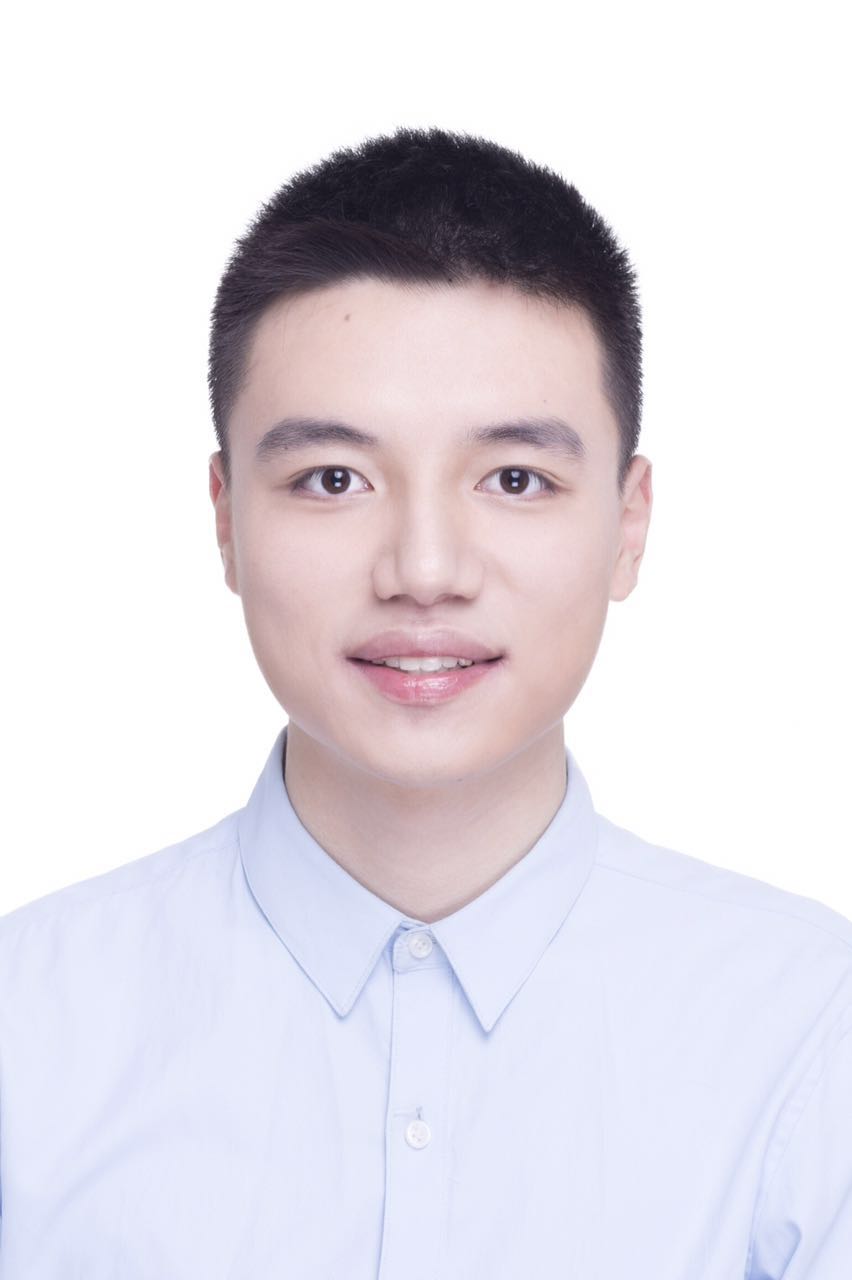}}]{Lumin Xu}
received the B.Eng. degree in information engineering from Zhejiang University, Hangzhou, China, in 2018. He is currently a Ph.D. candidate with the Department of Electronic Engineering, The Chinese University of Hong Kong, Hong Kong SAR, China. His research interests include computer vision, deep learning, and human pose estimation.
\end{IEEEbiography}

\begin{IEEEbiography}[{\includegraphics[width=1in,height=1.25in,clip,keepaspectratio]{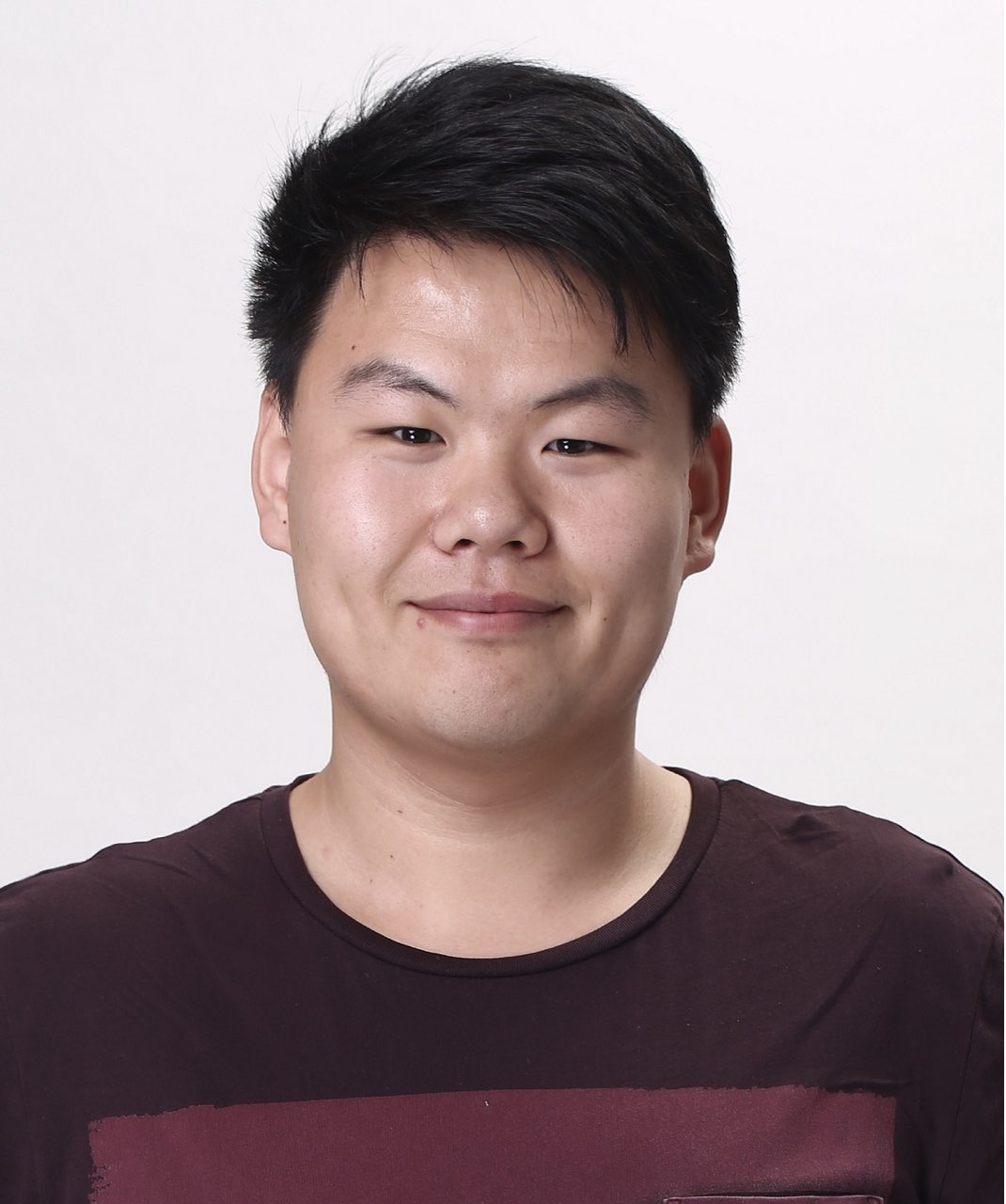}}]{Wentao Liu} received his Ph.D. degree in the School of EECS, Peking University. He is currently the Research Director of SenseTime, responsible for end-edge computing research. The research products are widely applied in augmented reality, smart industry, and business intelligence. His research interests include computer vision and pattern recognition.
\end{IEEEbiography}

\begin{IEEEbiography}[{\includegraphics[width=1in,height=1.25in,clip,keepaspectratio]{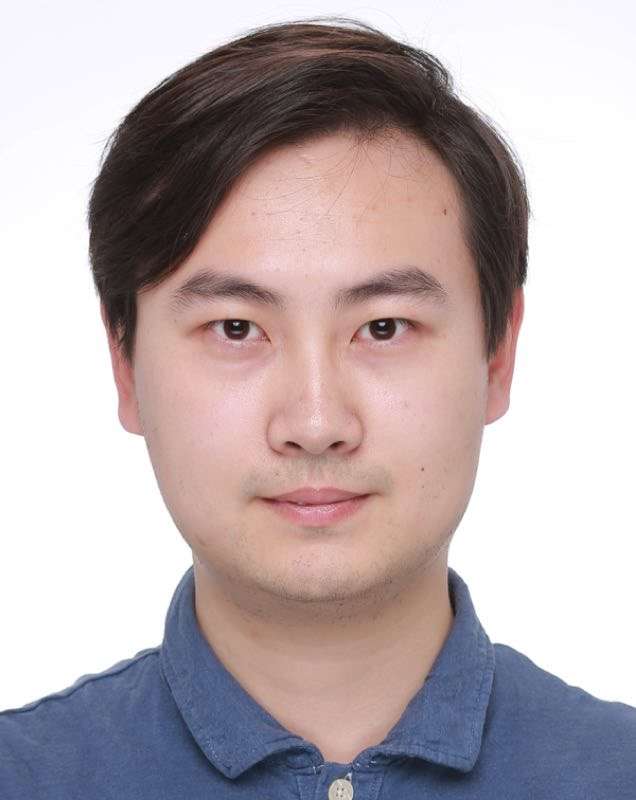}}]{Chen Qian} is currently the Executive Research Director of SenseTime, where he is responsible for leading the team in AI content generation and end-edge computing research in 2D and 3D scenarios. The technology is widely used in the top four mobile companies in China, APPs both home and abroad in augmented reality, video sharing and live streaming, vehicle OEMs, and smart industry. He has published dozens of articles on top journals and dozens of papers on top conferences, such as TPAMI, CVPR, ICCV, and ECCV with more than 4000 citations. He has also led the team to achieve the first place in the Competition of Face Identification and Face Verification in Megaface Challenge.
\end{IEEEbiography}

\begin{IEEEbiography}[{\includegraphics[width=1in,height=1.25in,clip,keepaspectratio]{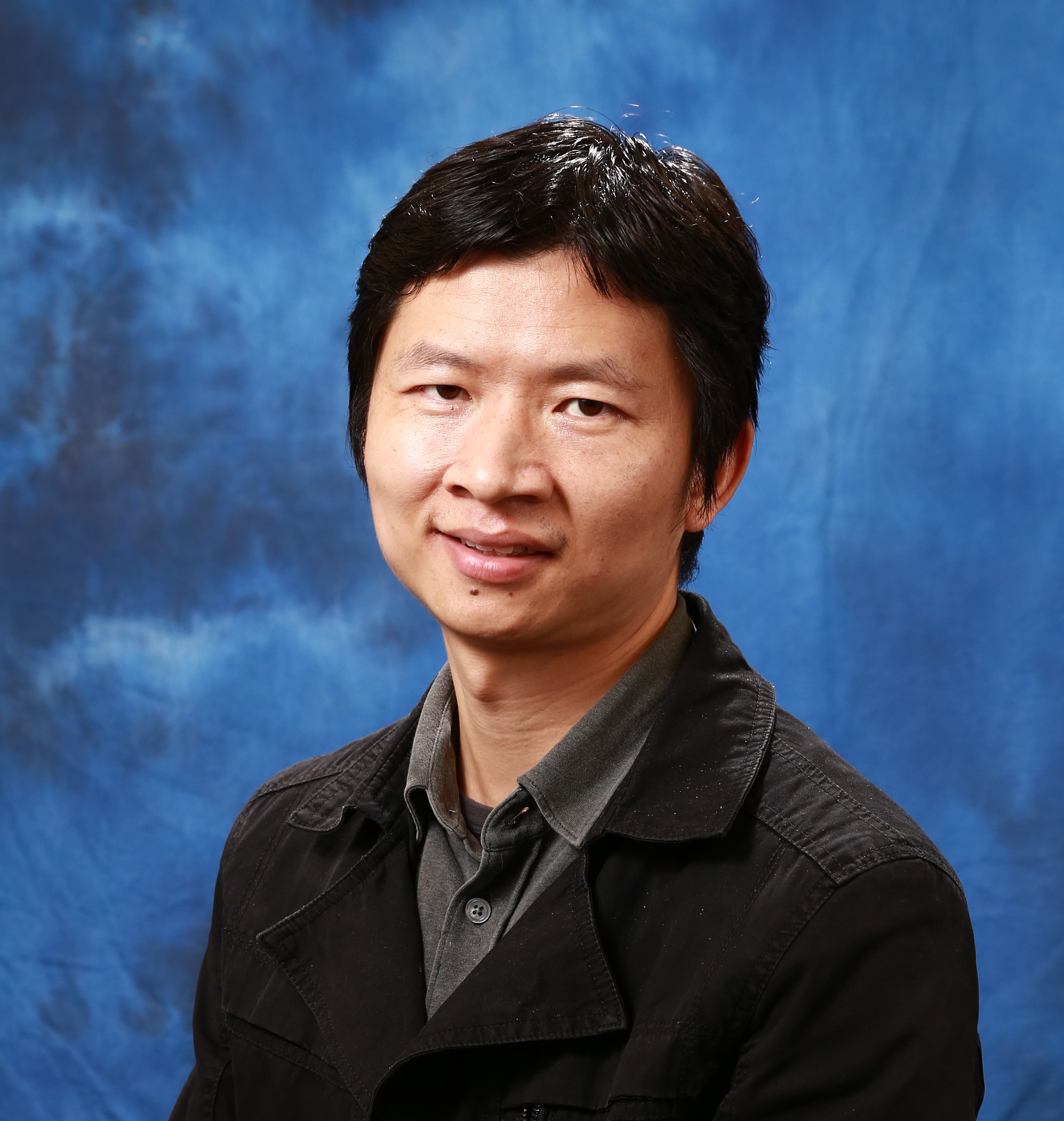}}]
{Wanli Ouyang} received the PhD degree in the Department of Electronic Engineering, The Chinese University of Hong Kong. He is now an associate professor in the School of Electrical and Information
Engineering at the University of Sydney, Australia. His research interests include image processing, computer vision and pattern recognition. He is a senior member of IEEE.
\end{IEEEbiography}

\begin{IEEEbiography}[{\includegraphics[width=1in,height=1.25in,clip,keepaspectratio]{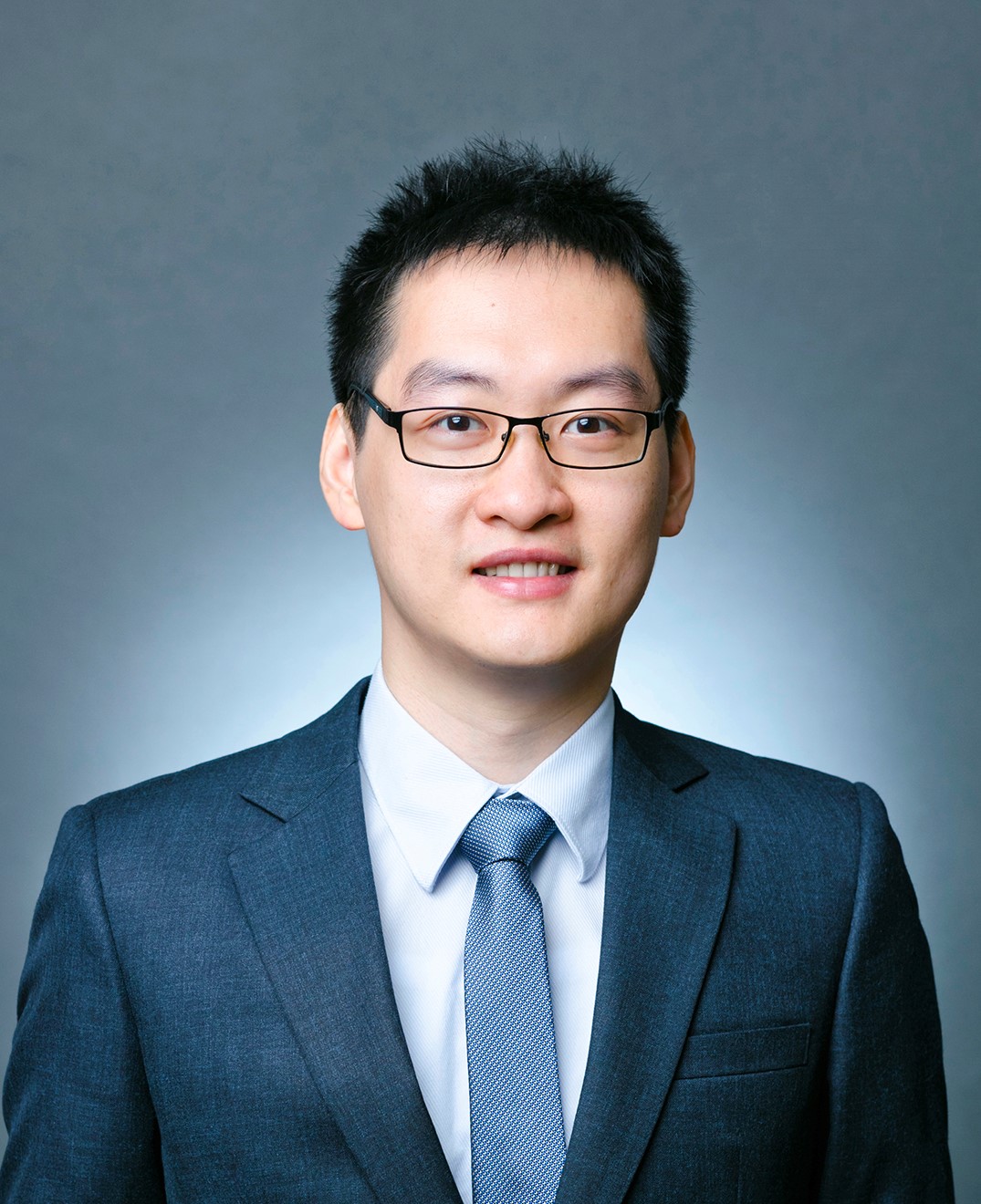}}]{Ping Luo}
is an Assistant Professor in the department of computer science, The University of Hong Kong (HKU). 
He received his PhD degree in 2014 from Information Engineering, the Chinese University of Hong Kong (CUHK), supervised by Prof. Xiaoou Tang and Prof. Xiaogang Wang. He was a Postdoctoral Fellow in CUHK from 2014 to 2016. 
He joined SenseTime Research as a Principal Research Scientist from 2017 to 2018. 
His research interests are machine learning and computer vision. He has published 100+ peer-reviewed articles in top-tier conferences and journals such as TPAMI, IJCV, ICML, ICLR, CVPR, and NIPS. His work has high impact with 18000+ citations according to Google Scholar. He has won a number of competitions and awards such as the first runner up in 2014 ImageNet ILSVRC Challenge, the first place in 2017 DAVIS Challenge on Video Object Segmentation, Gold medal in 2017 Youtube 8M Video Classification Challenge, the first place in 2018 Drivable Area Segmentation Challenge for Autonomous Driving, 2011 HK PhD Fellow Award, and 2013 Microsoft Research Fellow Award (ten PhDs in Asia).
\end{IEEEbiography}

\begin{IEEEbiography}[{\includegraphics[width=1in,height=1.25in]{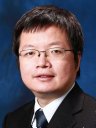}}]{Xiaogang Wang}
received the B.S. degree from the University of Science and Technology of China in 2001, the MS degree from The Chinese University of Hong Kong in 2003, and the PhD degree from the Computer Science and Artificial Intelligence Laboratory, Massachusetts Institute of Technology in 2009. He is currently a professor in the Department of Electronic Engineering at The Chinese University of Hong Kong. His research interests include computer vision and machine learning. 
\end{IEEEbiography}







%


\end{document}